\documentclass{article}

\usepackage{microtype}
\usepackage{graphicx}
\usepackage{subfigure}
\usepackage{booktabs} % for professional tables

\usepackage{chngcntr}
\usepackage{isomath}
\usepackage{mdwlist}
\usepackage{wrapfig}
\usepackage{amsmath,amssymb,amsthm}
\usepackage{amsfonts}
\usepackage{amsopn}
\usepackage{epsfig}
\usepackage{ifthen}
\usepackage{algorithm}
\usepackage{algorithmic}
\usepackage{float}
\usepackage{array}
\usepackage{multirow}
\usepackage{makecell}
\usepackage{caption}
\usepackage{multicol, lipsum}
\usepackage{array,colortbl,multirow,multicol,booktabs}

\newtheorem{theorem}{Theorem}

\newcommand{\obs}{\bm{z}}

\newcommand{\ft}{\bm{x}}
\newcommand{\ve}{\bm{u}}
\newcommand{\va}{\bm{a}}
\newcommand{\vh}{\bm{h}}
\newcommand{\vq}{\bm{q}}

\newcommand{\R}{{\rm I\!R}}

\newcommand{\N}{\mathbb{N}}

\newcommand{\E}{\text{\rm{I}\!\rm{E}}}

\usepackage{color}
\usepackage{hyperref}
\usepackage{pdfpages}
\usepackage{listings}
\usepackage{bm}

\usepackage{ctable}
\graphicspath{{figs/}}

\usepackage[accepted]{icml2019}

\icmltitlerunning{Deep Factors for Forecasting}
\begin{document}

\twocolumn[
\icmltitle{Deep Factors for Forecasting}



\icmlsetsymbol{equal}{*}

\begin{icmlauthorlist}
\icmlauthor{Yuyang Wang}{to}
\icmlauthor{Alex Smola}{to}
\icmlauthor{Danielle C. Maddix}{to}
\icmlauthor{Jan Gasthaus}{to}
\icmlauthor{Dean Foster}{to}
\icmlauthor{Tim Januschowski}{to}
\end{icmlauthorlist}

\icmlaffiliation{to}{Amazon Research}
% \icmlaffiliation{goo}{Googol ShallowMind, New London, Michigan, USA}
% \icmlaffiliation{ed}{School of Computation, University of Edenborrow, Edenborrow, United Kingdom}

\icmlcorrespondingauthor{Yuyang Wang}{yuyawang@amazon.com}
% \icmlcorrespondingauthor{Eee Pppp}{ep@eden.co.uk}

\icmlkeywords{Machine Learning, ICML}

\vskip 0.3in
]



\printAffiliationsAndNotice{} % otherwise use the standard text.

\begin{abstract}
Producing probabilistic forecasts for large collections of similar and/or dependent time series is a practically relevant and challenging task.
Classical time series models fail to capture complex patterns in the data,
and multivariate techniques struggle to scale to large problem sizes.  Their reliance on strong structural assumptions makes them data-efficient, and allows
them to provide uncertainty estimates.
The converse is true for models based on deep neural networks, which can
learn complex patterns and dependencies given enough data. In this
paper, we propose a hybrid model that incorporates the benefits of both
approaches. Our new method is data-driven and scalable via a latent, global,
deep component.  It also handles uncertainty through a local classical model. 
We provide both theoretical and empirical evidence for the
soundness of our approach through a necessary and sufficient
decomposition of exchangeable time series into a global and a 
local part.  Our experiments demonstrate 
the advantages of our model both in term of data efficiency, accuracy 
and computational complexity.
\end{abstract}

\section{Introduction}

Time series forecasting is a key ingredient in the automation and optimization of business processes.  In retail, decisions about which products to stock, when to (re)order them, and where to store them depend on forecasts of future demand in different regions; in (cloud) computing, the estimated future usage of services and infrastructure components guides capacity planning; regional forecasts of energy consumption are used to plan and optimize the generation of power; and workforce scheduling in warehouses and factories depends on forecasts of the future workload. 

%main idea: paragraph about switch from models designed by domain experts to data driven models
The prevalent forecasting methods in statistics and econometrics have
been developed in the context of forecasting individual or small groups of time
series.  The core of these methods is formed by comparatively simple (often linear) models,
which require manual feature engineering and model design by
domain experts to achieve good performance~\citep{harvey1990forecasting}. Recently, there has been
a paradigm shift from these model-based methods to fully-automated data-driven
approaches.  This shift can be attributed to the availability of large
and diverse time series datasets in a wide variety of fields, e.g.\
energy consumption of households, server load in a data
center, online user behavior, and demand for all products that a large
retailer offers.  These large datasets make it possible and necessary to learn models from data without significant manual work~\citep{bose2017probabilistic}. %~\citep{seeger2016bayesian}.  %Figure~\ref{fig:example} (Left) illustrates a typical large catalog forecasting problem.
% \begin{figure}[h]
%   \centering
%   \includegraphics[width=0.45\textwidth]{figs/expts}
%   %\includegraphics[width=0.3\textwidth]{figs/nyc_taxi_squish}
%   %\includegraphics[width=0.32\textwidth]{figs/taxi_fcst_123}
%   %\caption{Left: A typical forecasting problem in a large catalog~\citep{seeger2017approximate}. Middle: Spatio-temporal forecast problems for NYC taxi pickup. Right: An example of cold-start forecast (only using latitude and longitude with no observed time series information) with the proposed framework. 
%   \vspace{-.5cm}
%   \caption{A typical forecasting problem in a large catalog~\citep{seeger2017approximate}.}
%     \label{fig:example}
% \end{figure} 
%main message: how data can be used from related time series

A collection of time series can exhibit various dependency relationships between the individual time series that can be leveraged in forecasting.  These include: (1) local co-variate relationships (e.g.\ the price and demand for a product, which tend to be (negatively) correlated), (2) indirect relationships through shared latent causes (e.g.\ demand for multiple products increasing because an advertising campaign is driving traffic to the site), (3) subtle dependencies through smoothness, temporal dynamics, and noise characteristics of time series that are measurements of similar underlying phenomena (e.g.\ product sales time series tend to be similar to each other, but different from energy consumption time series).
The data in practical forecasting problems typically has all of these forms of dependencies. 
% A substantial amount of data consisting of past behavior of related
% time series can be leveraged for making a forecast for an individual
% time series.  For instance, this data could be energy
% consumption of other households, load of related servers, online
% behavior of similar users, and demand for similar products and
% covariates, such as price, holidays or promotions, respectively. 
Making use of this data from related time series allows more complex and potentially more accurate models to be fitted without overfitting.

Classical time series models have been extended to address the above dependencies of types (1) and (2) by allowing exogenous variables (e.g. the ARIMAX model and control inputs in linear dynamical systems), and employing multivariate time series models that impose a certain covariance structure (dynamic factor models), respectively. 
% TODO: add citation for ARIMAX and dynamic factor
Neural network-based models have been recently shown to excel in extracting complex patterns from large datasets of related time series by exploiting similar smoothness and temporal dynamics, and common responses to exogenous input, i.e.\ dependencies of type (1) and (3)~\citep{flunkert2017deepar, wen2017multi, mukherjee2018armdn,gasthaus2019probabilistic}.
These models struggle in producing calibrated uncertainty estimates.  They can also be sample-inefficient, and cannot handle type (2) dependencies. See ~\citep{faloutsos2018forecasting} for a recent survey on traditional and modern methods for forecasting.
%A commonality is that we have
%many related time series arising from the same context. In such scenarios, classical time series models such as Autoregressive Integrated
%Moving Average (ARIMA)~\citep{brockwell2013time} and exponential
%smoothing~\citep{hyndman2008forecasting} struggle to fit data well with limited per-time series information regime and fail to scale to large problems. This is where the neural network models shine, in particular, models based on RNNs~\citep{flunkert2017deepar,wen2017multi,mukherjee2018armdn} have shown to achieve better performance when facing the large catalog of related time series. However, lacking a principled way to deal with uncertainty, they are poor at estimating uncertainty of the predictive distributions. 

%\textcolor{red}{TODO: Add disadvantage of purely local and purely global methods}

%The challenges that need to be addressed are therefore the
%following. How can we build statistical models to efficiently and
%effectively learn to forecast from large and diverse data sources?
%How can we combine the data efficiency of the classical time series model and the expressive power that deep neural networks bring? 

The two main challenges that arise in the fully-automated data-driven approaches are: how to build statistical models that are able to borrow statistical strength and effectively learn to forecast from large and diverse data sources exhibiting all forms of dependencies, and how to combine the data efficiency and uncertainty characterization of classical time series models with the expressive power of deep neural networks.  In this paper, we propose a family of models that efficiently (in terms of sample complexity) and effectively (in terms of computational complexity) addresses these aforementioned challenges.

 %It can also help improve forecasts in the case of limited observations, known as cold-start problems.

% Renamed from literature review since it sounds very 'homework' like
\subsection{Background}
% here we are mixing a bit background and literature review; background is supposed to only introduce what things are, not about their advantages and disadvantages.}

%literature review of classical methods advantages
Classical time series models, such as general State-Space Models (SSMs), including ARIMA and exponential smoothing, excel at modeling the complex
dynamics of individual time series of sufficiently long history. For Gaussian State-Space Models, these methods are computationally efficient, e.g.\ via a Kalman filter, and provide uncertainty estimates. Uncertainty estimates are critical for optimal downstream decision making.  Gaussian Processes~\cite{rasmussen2006gaussian,seeger2004gaussian} (GPs) are another family of the models that have been applied to time series forecasting~\cite{girard2003gaussian,brahim2004gaussian}. These methods are
local, that is, they learn one model per time series.  As a
consequence, they cannot effectively extract information across
multiple time series. Finally, these classical methods struggle with cold-start problems, where more time series are added or removed over time. 

% GPs are Bayesian nonparametric methods, where one specifies a prior on a random function $f(\cdot)$ using a user-defined covariance structure $\mathcal{K}(\cdot, \cdot).$ 
% In addition, the classical time series methods normally require hand-crafted design for different time series. For example, SSM requires the user to specify the latent states such as level, trend and seasonality factors. For GPs, the design of an appropriate covariance function is crucial to produce accurate prediction. Thus, a large collection of time series makes it infeasible to tune to each one separately, making traditional models less appealing in such problems. 

% SSMs assume a (stochastic) latent state vector $\vh_t$ that evolves over time following a Markovian transition structure, $p(\vh_{t+1} | \vh_{<t+1}) = p(\vh_{t+1}|\vh_t).$ From the latent state, one generates the corresponding observations $z_t$ based on an emission model. 

Mixed effect models ~\citep{crawley2012mixed}  consist of two kinds of effects: fixed (global) effects that describe the whole population, and random (local) effects that capture the idiosyncratic of individuals or subgroups.  A similar mixed approach is used in Hierarchical Bayesian~\citep{gelman2013bayesian} methods, which combine global and local models to jointly model a population of related statistical problems. 
In~\citep{ahmed2012scalable,low2011multiple}, other combined local and global models are detailed.

%~\citep{ahmed2012scalable,low2011multiple}.  

Dynamic factor models (DFMs) have been studied in econometrics for decades to model the co-evolvement of multiple time series~\citep{geweke1977dynamic,stock2011dynamic,forni2000generalized,pan2008modelling}. DFMs can be thought as an extension of principal component analysis in the temporal setting.  All the time series are assumed to be driven by a small number of dynamic (latent) factors. Similar to other models in classical statistics, theories and techniques are developed with assuming 
 that the data is normally distributed and stationary. Desired theoretical properties are often lost when generalizing to other likelihoods. Closely related are the matrix factorization (MF) techniques~\citep{yu2016temporal,xie2017unified,hooi2019smf} and tensor factorization~\cite{araujo2019tensorcast}, which have been applied to the time series matrix with temporal regularization to ensure the regularity of the latent time series.  These methods are not probabilistic in nature, and cannot provide uncertainty estimation for non-Gaussian observations.

\subsection{Main Contributions}

In this paper, we propose a novel global-local method, Deep Factor
Models with Random Effects.  It is based on a global DNN backbone and
local probabilistic graphical models for computational efficiency.
The global-local structure extracts complex non-linear
patterns globally while capturing individual random effects for each
time series locally.

% We also extend the proposed frame
% to cover spatio-temporal forecasting scenarios.  The dynamical latent factors are extended 
% to temporal matrices, whose global part
% has a bilinear structure. Figure~\ref{fig:example} (center and
% right) illustrates an example of our framework applied to a spatio-temporal
% forecasting problem for NYC Taxi pickup. The right plot illustrates
% its generalization ability to previously unseen locations in a cold-start problem.

%Possibly too detailed for the intro: don't want to give too much of the model away early on sve for later
The main idea of our approach is to represent each time series,
or its latent function, as a combination of a global
time series and a corresponding local model. The global part is given
by a linear combination of a set of deep dynamic factors, where the
loading is temporally determined by attentions. The local model is
stochastic.  Typical local choices include white noise processes, linear
dynamical systems (LDS) or Gaussian processes (GPs). The stochastic
local component allows for the uncertainty to propagate forward in
time.  
%find appropriate location for this!!
%The variational Auto-Encoder (VAE)
%framework~\citep{kingma2013auto,rezende2014stochastic} enables
%efficient inference by decoupling the non-Gaussian emissions with the
%individual latent dynamics.
% 
%this method works for spatio-temporal forecasts show picture
%
Our contributions are as follows: i) Provide a unique characterization of exchangeable time series (Section~\ref{sec:problem}); ii) Propose a novel global-local framework for time series forecasting, based on i), that systematically marries deep neural networks and probabilistic models (Section~\ref{sec:model}); iii) Develop an efficient and scalable inference algorithm for non-Gaussian likelihoods that is generally applicable to any normally distributed probabilistic models, such as SSMs and GPs (Section~\ref{sec:model}). As a byproduct, we obtain new approximate inference methods for SSMs/GPs with non-Gaussian likelihoods; iv) Show how
state-of-the-art time series forecasting methods can be subsumed in
the proposed framework (Section~\ref{sec:rwork}); v) Demonstrate the accuracy and data efficiency of our approach through scientific experiments (Section~\ref{sec:exp}). %using RNNs and SSMs. 
\section{Exchangeable Series}
\label{sec:problem}
In this section, we formulate a general model for exchangeable time series.  A distribution over objects is exchangeable, 
if the probability of the objects is invariant under any %give mathematical definition with permuation?
permutation.  Exchangeable time series are a common occurrence.  For instance, user
purchase behavior is exchangeable, since there is no specific reason to assign a particular coordinate to a particular
user.  Other practical examples include sales statistics over similar products, prices of securities on the stock market and the use of
electricity. 
%necessary and sufficient decomp (unique characterization) of exchangeable time series 
\subsection{Characterization} %define notation more clearly

%Since a number of related sequence models exist, it is only natural to
%ask whether there exists a characterization of models describing
%exchangeable time series. 
Let $\obs_i \in \mathcal{Z}^T$, where
$\obs_i$ denotes the $i^{\text{th}}$ exchangeable time series, $\mathcal{Z}$ denotes the domain of
observations and $T \in \mathbb{N}$ denotes the length of the time
series.\footnote{Without loss of generality and to avoid notational
  clutter, we omit the extension to time series beginning at different
  points of time. Our approach is general and covers these cases as
  well.}  We denote individual observations at some time $t$ as
$z_{i,t}$.  We assume that we observe $z_{i,t}$ at discrete time
steps %for now assumption
to have a proper time series rather than a marked point process.

\begin{theorem}
  \label{th:definetti-time}
  Let $p$ be a distribution over exchangeable time series $\obs_i$
  over $\mathcal{Z}$ with length $T$, where $1 \leq i \leq N$. Then %1<= t <= T
  $p$ admits the form
  \begin{align*}
    p(z) = \int \prod_{t=1}^T p(g_t|g_{1:t-1}) \prod_{i=1}^N
    p(z_{i,t}|z_{i,1:t-1}, g_t) dg.
  \end{align*}
\end{theorem}
In other words, $p(z)$ decomposes into a global time series $g$ and
$N$ local times series $\obs_i$, which are conditionally
  independent given the latent series $g$. 
\begin{proof}
It follows from de Finetti's theorem~\citep{diaconis1977_definetti, diaconis1980_definetti} that 
\begin{equation}
	p(z) = \int p(g) \prod_{i=1}^N p(\obs_i|g) dg.
	\label{eqn:definetti}
\end{equation} 
Since $\obs_i$ are time series, we can decompose $p(\obs_i|g)$ in
the causal direction using the chain rule as $$p(\obs_i|g) = \prod_{t=1}^T p(z_{i,
  t}|z_{i,1:t-1},g).$$ Substituting this into the de Finetti factorization in Eqn. \eqref{eqn:definetti}
gives 
\begin{align*}
  p(z) = \int p(g) \prod_{i=1}^N \prod_{t=1}^T 
  p(z_{i,t}| z_{i,1:t-1},g) dg.
\end{align*}
Lastly, we can decompose $g$, such that $g_t$ contains a
sufficient statistic of $g$ with respect to $z_{\cdot,t}$.  This holds
trivially by setting $g_t = g$, but defeats the purpose of
the subsequent models. Using the chain rule on $p(g)$ and substituting the result in proves the claim. 
\end{proof}

\begin{theorem}
  For tree-wise exchangeable time series, that is time series that can be grouped hierarchically into exchangeable sets, there exists a corresponding set of hierarchical latent variable
  models. 
\end{theorem}
The proof is analogous to that of Theorem \ref{th:definetti-time}, and follows from a hierarchical
extension of de Finetti's theorem \citep{austin2014hierarchical}. 
This decomposition is useful when dealing with product
hierarchies. For instance, the sales events within the category of
iPhone charger cables and the category of electronics cables may be exchangeable. 

%A hierarchical version of the de Finetti and Aldous-Hoover representations
%Article in Probability Theory and Related Fields 159(3-4) · January 2013

%why under exchangeable time series section??
\subsection{Practical Considerations}
\label{sec:practical}
We now review some common design decisions used in modeling time series. The first is to replace
the decomposition $\prod_{t=1}^T p(z_{i,t}|\bm{z}_{i,1:t-1})$ by a tractable, approximate statistic $h_t$ of the past, such
that $p(z_{i,t}|\bm{z}_{i,1:t-1}) \approx p(z_{i,t}|h_{i,t})$. Here, $h_t$ typically
assumes the form of a latent variable model via
$p(h_{i,t}|h_{i,t-1}, z_{i,t-1})$. Popular choices for real-valued random
variables are SSMs and GPs. 

The second is to assume that the global variable $g_t$ is drawn from some
$p(g_t|g_{t-1})$.  The inference in this model is costly, since it requires constant interaction, via Sequential
Monte Carlo, variational inference or a similar procedure between
local and global variables at prediction time. One way to reduce these expensive calculations is to incorporate past local
observations $z_{\cdot,t-1}$ explicitly. While this somewhat negates
the simplicity of Theorem~\ref{th:definetti-time}, it yields
significantly higher accuracy for a limited computational
budget, 
$g_t \sim p(g_t| g_{t-1}, z_{\cdot,t-1}).$

Lastly, the time series often comes with observed covariates, such as a user's location or a detailed
description of an item being sold.  We add these
covariates $\ft_{i,t}$ to the time series signal to obtain the following model:
\begin{align}
  \label{eq:practical-model}
  p(z|x) = \int &\prod_{t=1}^T p(g_t| g_{t-1}, x_{\cdot,t}, z_{\cdot,t-1})\hspace{.1cm} \times  \\
  &\prod_{i=1}^N \Big[ p(h_{i,t}|g_t, h_{i,t-1}, z_{i,t-1}, \ft_{i,t}) \hspace{.1cm}  \times  \nonumber\\ 
  &p(z_{i,t}|g_t, h_{i,t}, z_{i,t-1}, \ft_{i,t}) \Big] dg dh. \nonumber
\end{align}
Even though this model is rarely used in its full generality, Eqn.
(\ref{eq:practical-model}) is relevant because it is by some measure the
most general model to consider, based on the
de Finetti factorization in Theorem~\ref{th:definetti-time}. 

%move related work to literature review in intro.  we should not go backwards here.  the paper should now flow into special cases of the model.
\subsection{Special Cases} %not sure related work should be in this section or in intro

% Let $\mathcal{X}\subset \R^d$ denote the input features space and $\mathcal{Z}\subset \R^k$ the space of the observations. We are given a set of $N$ time series with the $i$-th time series consisting of $(\ft_{i,t}, z_{i,t})\in \mathcal{X} \times \mathcal{Z}, t=1, \cdots, T_i,$ where $\ft_{i,t}$ are the input co-variates and $z_{i,t}$ is the corresponding observation at time $t$. Given a forecast horizon $\tau_i \in \mathbb{N}^+$, our goal is to calculate the joint distribution of future observations, 
% \begin{equation}
% \Pr(\{\obs_{i, T_i+1:T_i+\tau_i}\}_{i=1}^N | \{\ft_{i, T_i+1:T_i+\tau_i}\}_{i=1}^N, \{\mathcal{D}\}_{i=1}^N), 
% \end{equation}
% where $\mathcal{D}_i = \{(\ft_{1:T_i}, \obs_{1:T_i})\}$ denotes the $i$-th time series with corresponding features. As the reader have realdy noticed, here we use the shorthand notation $1:T$ to denote $\{1, 2, \cdots, T\}.$ For the sake of concreteness, we restrict ourselves to univariate time series ($k=1$). %We drop the index of the time series $i$ whenever it is implied by the context. 
%The dynamic factors are normally taken to be the lagged values of the time series. 

The global-local structure has been used previously in a number of
special contexts~\citep{xu2009multi,ahmadi2011multi,hwang2016automatic,choi2011lifted}. For instance, in Temporal LDA
\citep{ahmed2012scalable} we assume that we have a common
fixed Dirichlet-Multinomial distribution capturing the
distribution of tokens per topic, and a time-variant set of latent
random variables capturing the changes in user preferences. This
is a special case of the above model, where the global time
series does not depend on time, but is stationary instead. 

A more closely related case is the Neural Survival Recommender model of
\citep{jing2017neural}. This models the temporal dynamics of
return times of a user to an app via survival analysis. In particular,
it uses a LSTM for the global dynamics and LSTMs for the local
survival probabilities. In this form, it falls into the
category of models described by
Theorem~\ref{th:definetti-time}. Unlike the models we propose in this
paper, it does not capture local uncertainty accurately. It also primarily deals with point processes rather than proper time
series, and the inference algorithm differs quite significantly.

\section{Deep Factor Models with Random Effects}
\label{sec:model}
 %\todo{this should be expanded and maybe move to elsewhere. }
% Inspired by the structure time series models~\citep{harvey1990forecasting}, dynamic factor models~\citep{geweke1977dynamic}, as well as recently development in RNN-based forecasters~\citep{flunkert2017deepar}, 

% \onecolumn
% \begin{center}
% \begin{tabular}{rl}
% \hline
% \textsc{notation} & \textsc{comment}\\
% \hline
% $i$ & index of the time series \\
% $t$ & index of time\\
% $z_{i,t} \in \R$ &  value of the $i$-th time series at time $t$\\
% $T_i \in \N^+$ &  length of the $i$-th time series\\
% $u_{i,t} \in \R$ &  latent function values of the $i$-th time series at time $t$\\
% $\ft_{i,t}\in \R^d$ &  input feature of the $i$-th time series at time $t$\\
% $g_t(\cdot) \in \R^K$ &  output of the global model at time $t$ \\
% $w_{i,t}(\cdot) \in \R^K$ &  output of the attention model at time $t$ for the $i$-th time series\\
% $r_{i,t}(\cdot) \in \R$ &  output of the local model $\mathcal{R}$ at time $t$ \\
% $h_{i,t} \in \R^l$ &  latent state vector in the local probabilistic model (if available) \\
% \hline
% \end{tabular}
% \end{center}

%\todo[inline]{A better motivation in the beginning, especially regarding the connection to the dynamic factor model and mixed-effects model.}

Motivated by the characterization of exchangeable time series, in this section, we propose a general framework for global-local
forecasting models, called Deep Factor Models with Random Effects, that follows the structure given by the decomposition in Theorem~\ref{th:definetti-time}. 
We describe the family of the methods, show three concrete instantiations (DF-RNN, DF-LDS, and DF-GP), and derive the general inference and learning algorithm.  Further models that can be obtained within the same framework, and additional details about the design choices, are described in Appendix~\ref{seq:dfm}.

%\subsection{Problem Definition} 
%Let, $\mathcal{X}\subset \R^d$ denote the input features space and $\mathcal{Z}\subset \R^k$ the space of the observations. 
We are given a set of $N$ time series, with the $i^{\text{th}}$ time series consisting of tuples $(\ft_{i,t}, z_{i,t})\in \mathbb{R}^d \times \mathbb{R}, t=1, \cdots, T,$ where $\ft_{i,t}$ are the input co-variates, and $z_{i,t}$ is the corresponding observation at time $t$.  Given a forecast horizon $\tau \in \mathbb{N}^+$, our goal is to calculate the joint predictive distribution of future observations, 
%% to be added in.
 \[
 p(\{z_{i, T+1:T+\tau}\}_{i=1}^N  | \{\ft_{i, 1:T+\tau}, z_{i, 1:T}\}_{i=1}^N),
 \]
 i.e.\ the joint distribution over future observations given all co-variates (including future ones) and past observations. 

 %For concreteness, we restrict ourselves to the case of univariate observations ($k=1$). 
% The probabilistic predictions, especially for forecasting, is crucial for making statistical decisions for the downstream system. %We use the notation $1:T$ from the previous section to denote $\{1, 2, \cdots, T\}.$  %We drop the index of the time series $i$ whenever it is implied by the context. 

\subsection{Generative Model}
\label{sect:genmodel}

%\vspace{-1cm}
Our key modeling assumption is that each time series $\obs_{i,t}, t=1, 2, \ldots$ is governed by a fixed global (non-random) and a random component, whose prior is specified by a generative model $\mathcal{R}_i$. 
%The observed value $z_{i,t}$ at time $t$ or more generally, its latent function $u_{i}(\cdot)$ can be expressed as a sum of
%the weighted average of the global patterns and its local fluctuations, and the generative process reads
In particular, we assume the following generative process:
\begin{align}
\text{global factors}:\quad & g_k(\cdot) = \textsc{rnn}_k(\cdot), \quad k = 1, \cdots, K,\nonumber\\
\text{fixed effect}: \quad & f_{i}(\cdot) = \sum_{k=1}^K w_{i,k}\cdot g_k(\cdot),\label{eqn:fixed}\\
\text{random effect}:\quad &r_{i}(\cdot) \sim \mathcal{R}_i,\quad  i = 1, \cdots, N,\label{eqn:random}\\
\text{latent function}:\quad  &u_i(\cdot) = f_{i}(\cdot) + r_i(\cdot),\label{eqn:u}\\
\text{emission}: \quad & z_{i,t} \sim p(z_{i,t}|u_i(\ft_{i,t})), \nonumber 
\end{align}
The observation model $p$ can be any parametric distribution, such as Gaussian, Poisson or Negative Binomial. All the functions $g_k(\cdot), r_i(\cdot), u_i(\cdot)$ take features $\ft_{i,t}$ as input, and we define $u_{i,t} := u_i(\ft_{i,t})$, the embedding $\bm{w}_i := [w_{i,k}]_k.$ % and $\bm{g}_t = [g_k(\ft_t)]_k$.

\begin{figure}[h]
  \begin{center}
    \includegraphics[width=.25\textwidth]{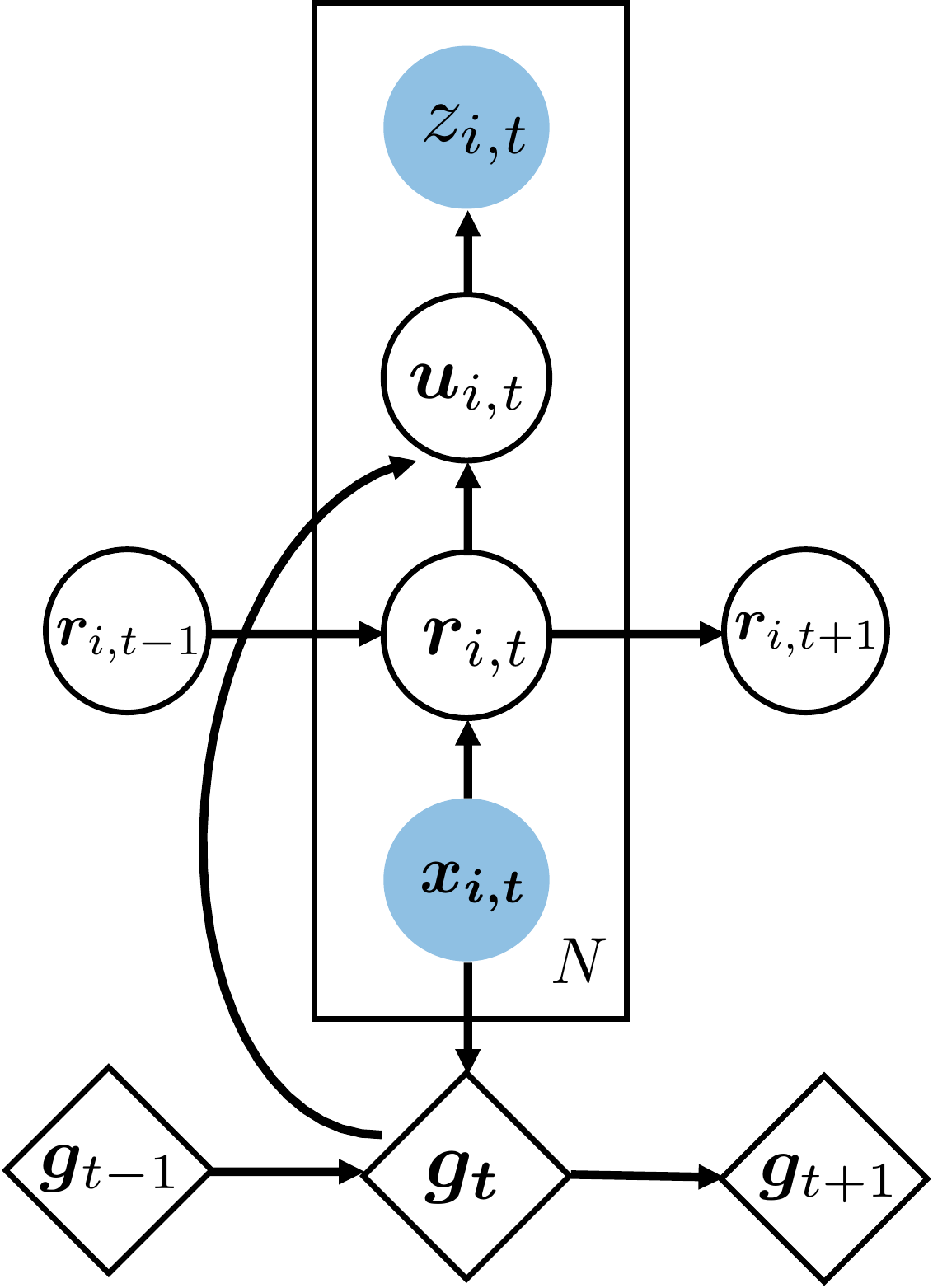}
  \end{center}
  \caption{Plate graph of the proposed Deep Factor Model with Random Effects. The diamond nodes represent deterministic states.}
  \label{fig:platemodel}
  \vspace{-.5cm}
\end{figure}

%\todo{describe the item features and item dynamic features.}

\begin{table*}[ht]
\centering
\begin{tabular}{l|m{4cm}|c|ccc}
\toprule
\textsc{name} &  \textsc{description} & \textsc{local} & \textsc{likelihood (gaussian case)}\\
\midrule
DF-RNN & Zero-mean Gaussian noise process given by \textsc{rnn} & $r_{i,t} \sim \mathcal{N}(0, \sigma_{i,t}^2)$ & $p(\obs_i) = \prod_{t}\mathcal{N}(z_{i,t} - f_{i,t} | 0, \sigma_{i,t}^2)$\\
\midrule
DF-LDS & State-space models & $r_{i,t} \sim \text{LDS}_{i,t}$ (cf. Eqn. \eqref{eqn:df_issm}) & $p(\obs_i)$ given by Kalman Filter\\
\midrule
DF-GP & Zero-mean Gaussian Process  & $r_{i,t}\sim \text{GP}_{i,t}$ (cf. Eqn. \eqref{eqn:df_gp})& $p(\obs_i) = \mathcal{N}(\obs_i -  \bm{f}_i| \bm{0}, \mathbf{K}_{i} + \bm{\sigma}_i^2\textbf{I})$\\
\bottomrule
\end{tabular}
%\caption{Summary of the datasets used in this paper. Both DF-RNN and TDF-RNN are described in Section~\ref{sec:df-rnn}.}
\caption{Summary table of Deep Factor Models with Random Effects. The likelihood column is under the assumption of Gaussian noise. }%The notation $\ominus$ means either subtraction or division.}
\label{tab:model_sums}
\end{table*}

\subsubsection{Global effects (common patterns)} 
The global effects are given by linear combinations of $K$ latent 
global deep factors modeled by RNNs. These deep factors can be thought of as dynamic principal components or eigen time series that drive the underlying dynamics of all the time series.  As mentioned in Section~\ref{sec:practical}, we restrict the global effects to be deterministic to avoid costly inference at the global level that depends on all time series.

%The global deep factors takes time features $\bm{x}_t$ that are independent of each time series. 

% For each time series at time $t$, we use attention networks to assign attentions $w_{i,t} \in\R^K$ to the dynamic factors $g_t$.  This determines the group of the global factors to focus on and the relevant segment of histories. At a high level, the weighting gives temporal attention to different global factors. % (dynamic loadings in DFM language). 

%\textbf{Local fluctuations (random effect):} 

\begin{figure}[H]
    \centering
    \includegraphics[width=.79\linewidth]{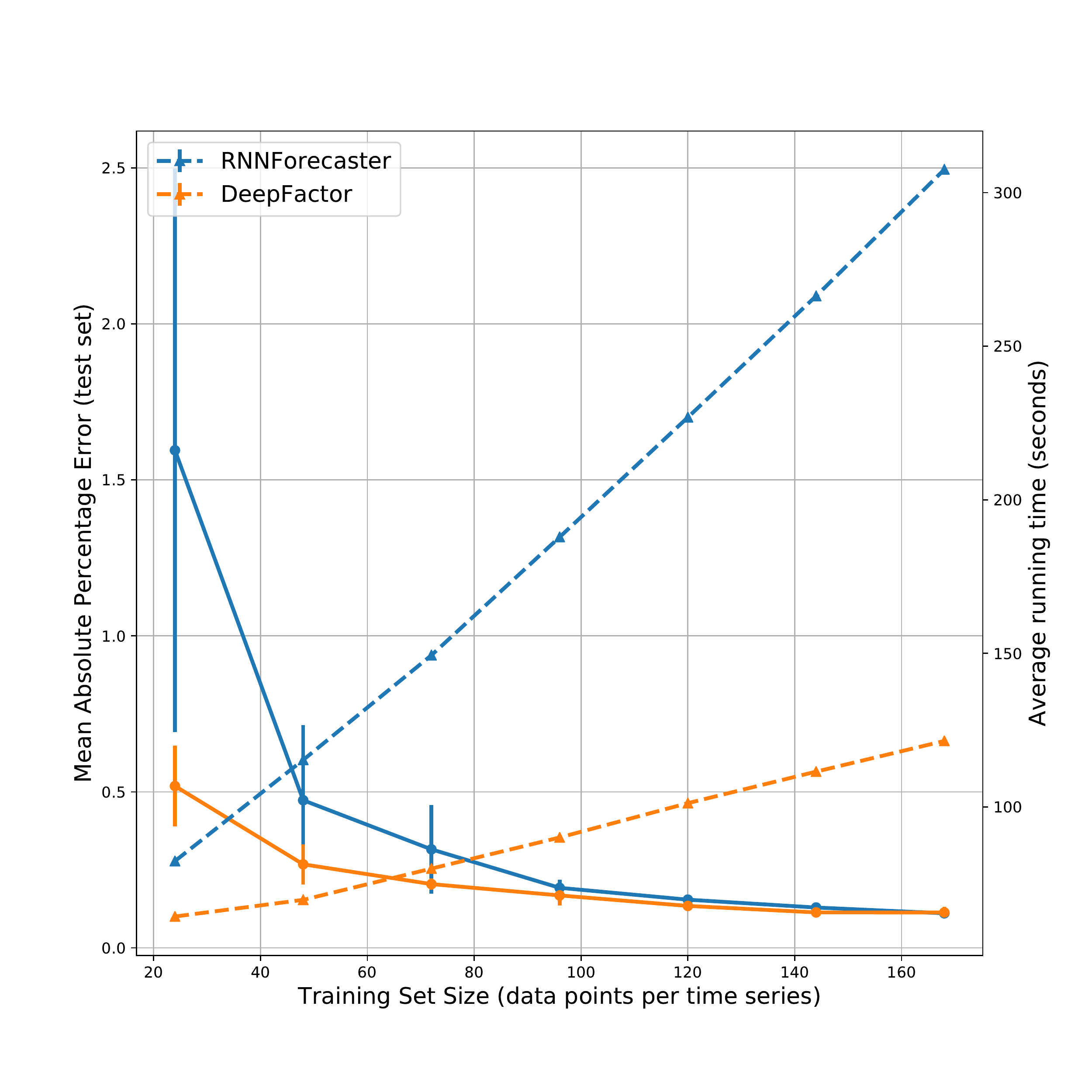} 
    \vspace{-.5cm}
    \caption{Generalization Error (solid line), Mean Absolute Percentage Error (MAPE) on the test set and running time in seconds (dashed line) vs. the size of the training set, in terms of data points per time series. The experiments are repeated over 10 runs.}
    \label{fig:sc}
\end{figure}

%We would like to remark that the novel usage of RNN in the form (\ref{eqn:fixed}) have its merit by itself. 
The novel formulation of the fixed effects from the RNN in Eqn. \eqref{eqn:fixed} has advantages in comparison to a standard RNN forecaster.  Figure~\ref{fig:sc} compares the generalization errors and average running times of using Eqn. \eqref{eqn:fixed} with the $L_2$ loss and a standard RNN forecaster with the same $L_2$ loss and a comparable number of parameters on a real-word dataset \texttt{electricity}~\cite{Dua:2017}.  Our fixed effect formulation shows significant data and computational efficiency improvement.  The proposed model has less variance in comparison to the standard structure. Detailed empirical explorations of the proposed structures can be found in Section~\ref{sec:synthetic}.
\vspace{-.5cm}

\subsubsection{Random effects (local fluctuations)} 
The random effects $r_{i}(\cdot)$ in Eqn. \eqref{eqn:random} can be chosen to be any classical probabilistic time series model $\mathcal{R}_i$.  To efficiently compute their marginal 
likelihood $p(\obs_i | \mathcal{R}_i)$, $r_{i,t}$ should be chosen to satisfy the normal distributed observation assumption. Table~\ref{tab:model_sums} summarizes the three models we consider for the local random effects models $\mathcal{R}_i$.

The first local model, DF-RNN, is defined as $r_{i,t} \sim \mathcal{N}(0, \sigma_{i,t}^2)$, where $\sigma_{i,t}$ is given by a noise RNN that takes input feature $\ft_{i,t}$. The noise process becomes correlated with the covariance function implicitly defined by the RNN, resulting in a simple deep generative model.

The second local model, DF-LDS, is a part of a special and robust family of SSMs, Innovation State-Space Models (ISSMs)~\citep{hyndman2008forecasting, seeger2016bayesian}. %ISSMs are robust with solid performance in a wide variety applications~\citep{seeger2016bayesian}.  
This gives the following generative model:
\begin{equation}
\begin{split}
\vh_{i,t} &= \mathbf{F}_{i,t}\vh_{i,t-1} + \vq_{i,t} \epsilon_{i,t}, \quad \epsilon_{i,t} \sim \mathcal{N}(0, 1), \\
r_{i,t} &= \va_{i,t}^\top \vh_{i,t}. %+ \nu_{i,t}.%\quad \nu_{i,t} \sim \mathcal{N}(0, \sigma^2_{i}). %, \quad \epsilon_{i,t} \sim \mathcal{N}(0, 1)  \\
%& \epsilon_{i,t} \sim \mathcal{N}(0, 1),\quad \nu_{i,t} \sim \mathcal{N}(0, \sigma^2_{i,t}).
% \text{fixed effect}:\quad  &f_{i,t} = w_i^\top g_t(\ft_{i,t}),\\
% \text{emission}: \quad & z_{i,t} \sim p(\cdot|u_{i,t}),\ u_{i,t} = f_{i,t} + r_{i,t}.
\label{eqn:df_issm}
\end{split}
\end{equation}
The latent state $\vh_{i,t}$ contains information about the level, trend, and seasonality patterns.  It evolves by a deterministic transition matrix $\mathbf{F}_{i,t}$ and a random innovation $\vq_{i,t}.$ The structure of the transition matrix $\mathbf{F}_{i,t}$ and innovation strength $\vq_{i,t}$ determines which kinds of time series patterns the latent state $\vh_{i,t}$ encodes (cf. Appendix~\ref{sec:df-lds} for the concrete choice of ISSM). 

The third local model, DF-GP, is defined as the Gaussian Process, 
\begin{equation}
	r_{i,t}\sim \text{GP}(0, \mathcal{K}_i(\cdot, \cdot)),  %+ \sigma_{i,t}^2\cdot\delta(\cdot, \cdot)),
	\label{eqn:df_gp}
\end{equation}
 where $\mathcal{K}_i(\cdot, \cdot)$ denotes the kernel function.  In this model, with each time series has its own set of GP hyperparameters to be learned. % We use square exponential kernels. 

% shall we add? 
% \paragraph{Extensions and Special Cases}
% The flexibility of the proposed framework lies in the fact that it nicely interpolates between purely local to purely global and subsequent subsumed a variety of special cases. For example, if one chooses only one factor with no random effects, accompanied with autoregressive inputs, we arrive at DeepAR~\cite{flunkert2017deepar}. Furthermore, in our formation, the scale of each time series is automatically estimated rather than pre-specified as in DeepAR. Similarly, changing the emission probability to Gaussian Mixtures leads us to AR-MDN~\cite{mukherjee2018armdn}. On the other hand, a family of models that cannot be subsumed into our framework is the Sequence-to-Sequence models for forecasting~\citep{wen2017multi}, which make predictions discriminatingly rather than generatively. The proposed framework, with a small addition, is readily applicable for spatio-temporal forecasting, See Appendix~\ref{app:spatio}) for detail.

\subsection{Inference and Learning}
\label{sect:inference}
Given a set of $N$ time series, our goal is to jointly estimate $\mathbf{\Theta}$, the parameters in the global RNNs, the embeddings and the hyper-parameters in the local models. We use maximum likelihood estimation, where
$
\mathbf{\Theta} = \text{argmax} \sum_i\log p(\obs_i | \mathbf{\Theta}). 
$
Computing the marginal likelihood may require doing inference over the latent variables. The general learning algorithm is summarized in Algorithm~\ref{alg:1}.

\begin{algorithm}[htp]
    \caption{Training Procedure for Deep Factor Models with Random Effects.}
    \label{alg:1}
    \begin{algorithmic}[1] % The number tells where the line numbering should start
    \FOR{each time series $\{(\ft_{i}, \obs_{i})\}$} %, \quad t = 1:T$ in a mini-batch}
      %\FOR{each time series $\{(\ft_{i,t}, z_{i,t})\}, \quad t = 1:T$ in a mini-batch}
        \STATE Sample the estimated latent representation from the variational encoder $\widetilde{\ve}_i \sim q_\phi(\cdot | \obs_i) $ for non-Gaussian likelihood, otherwise $\widetilde{\ve}_i := \obs_i.$ % i.e., 
        % \[
        %   \widetilde{u}_{i,t} \sim \prod_{t=1}^{T_i} q_\phi(\widetilde{u}_{i,t} | \obs_i), 
        % \]        
        \STATE With the current estimate of the model parameters $\mathbf{\Theta},$ compute the fixed effect 
        $
        f_{i,t} = \sum_{k=1}^K w_{i,k}\cdot g_k(\ft_{i,t}), 
        $
        and corresponding ISSM parameters for DF-LDS or the kernel matrix $\mathbf{K}_i$ for DF-GP.
      %where $\ominus$ means either subtraction or division.    
    \STATE Calculate the marginal likelihood $p(\obs_i)$ as in Table~\ref{tab:model_sums} or its variational lower bound as in Eqn. \eqref{eqn:selbo}. % for each time series given the estimate of the local effects.
    \ENDFOR
    \STATE Accumulate the loss in the current mini-batch, and perform stochastic gradient descent. %with respect to $\bm{\mathbf{\Theta}}$ and $\phi$ (if needed), the parameters in global factor, attention networks, and recognition network (if needed).
    \end{algorithmic}       
\end{algorithm}

%For the ease of exposition, we drop the index of the time series $i$ whenever the context is clear. 
\subsubsection{Gaussian Likelihood} 
\label{sec:gaussian}
When the observation model $p(\cdot|u_{i,t})$ is Gaussian, $z_{i,t}\sim \mathcal{N}(u_{i,t}, \sigma_{i,t}^2)$, the marginal likelihood can be easily computed for all three models. Evaluating the marginal likelihood for DF-RNN is straightforward (see Table \ref{tab:model_sums}). 

For DF-LDS and DF-GP, the Gaussian noise can be absorbed into the local model, yielding 
$
	z_{i,t} = u_{i,t} = f_{i,t} + r_{i,t},
$
where $r_{i,t}$, instead of coming from the noiseless LDS and GP, is generated by the noisy versions. More precisely, for DF-LDS, $r_{i,t} = \va_{i,t}^\top \vh_{i,t}+ \nu_{i,t}$ and $\nu_{i,t} \sim \mathcal{N}(0, \sigma^2_{i})$, and the marginal likelihood is obtained with a Kalman filter. In DF-GP, it amounts to adding $\sigma_{i,t}^2\cdot\delta(\cdot, \cdot)$ to Eqn~\eqref{eqn:df_gp}, where $\delta(\cdot, \cdot)$ is the Dirac delta function. The marginal likelihood becomes the standard GP marginal likelihood, which is the multivariate normal with mean $\bm{f}_i := [f_i(\ft_{i,t})]_t$ and covariance matrix $\mathbf{K}_i + \bm{\sigma_i}^2\mathbf{I}$, where $\mathbf{K}_i := [\mathcal{K}(\ft_{i,t}, \ft_{i, t})]_t$ and $\mathbf{I}$ is the identity matrix of suitable size.

\subsubsection{Non-Gaussian Likelihood}
\label{subsect:nongauss}

When the likelihood is not Gaussian, the exact marginal likelihood is intractable.  We use variational inference, and optimize a variational lower bound of the marginal likelihood $\log p(\obs)$: 
\begin{equation}
\log \int p(\obs, \ve, \vh)   \geqslant \E_{q_\phi(\ve, \vh)}\log\left[\frac{p(\obs, \ve, \vh)}{q_\phi(\ve,\vh)} \right], 
\label{eq:bound}
\end{equation}
where $\ve$ is the latent function values, and $\vh$ is the latent states in the local probabilistic models \footnote{For cleaner notation, we omit the time series index $i$}.  Optimizing this stochastic variational lower bound for the joint model over all time series is computationally expensive.  

We propose a new method that leverages the structure of the local probabilistic model to enable fast inference at the per-time series level. This enables parallelism and efficient inference that scales up to a large collection (millions) of time series.  Motivated by~\citep{fraccaro2017disentangled}, we choose the following structural approximation $q_\phi(\vh, \ve|\obs) := q_\phi(\ve|\obs)p(\vh|\ve),$
where the second term matches the exact conditional posterior with the random effect probabilistic model $\mathcal{R}$. With this form, given $\ve$, the inference becomes the canonical inference problem with $\mathcal{R}$, from Section \ref{sec:gaussian}.  The first term $q_\phi(\ve|\obs)$ is given by another neural network parameterized by $\phi$, called a recognition network in the variational Auto-Encoder (VAE) framework~\citep{kingma2013auto,rezende2014stochastic}. After massaging the equations, the stochastic variational lower bound in Eqn. \eqref{eq:bound} becomes 
\begin{align}
&\E_{q_\phi(\ve)}\left[\log\left[\frac{p(\obs|\ve)}{q_\phi(\ve|\obs)}\right] + \log p(\ve)\right]\label{eqn:elbo}\\
&\ \approx\frac{1}{L}\left(\log p(\obs|\widetilde{\ve}_j) + \log p(\widetilde{\ve}_j) - \log q_\phi(\widetilde{\ve}_j|\obs)\right),\label{eqn:selbo}
\end{align}
with $\widetilde{\ve}_j \sim q_\phi(\ve)$ for $j=1, \cdots, L$ sampled from the recognition network. The first and third terms in Eqn. \eqref{eqn:selbo} are straightforward to compute. For the second term, we drop the sample index $j$ to obtain the marginal likelihood $\log p(\widetilde{\ve})$ under the normally distributed random effect models.  This term is computed in the same manner as in Section~\ref{sec:gaussian}, with $\obs_i$ substituted by $\widetilde{\ve}$.  When the likelihood is Gaussian, the latent function values $\bm{u}$ are equal to $\bm{z},$ and we arrive at $\log p(\obs)$ from Eqn. \eqref{eqn:elbo}.

\section{Related Work and Discussions}
\label{sec:rwork}
%\todo{discussion with several models.}

% In the early nineties, Feedforward NNs were popular among forecasters~\cite{zhang1998forecasting} with applications in electrical load~\cite{park1991electric,lu1993neural}, financial time series~\cite{gately1995neural}, and others~\cite{hill1994artificial}. 
% Recent ground-breaking successes of deep neural network in other areas of machine learning have brought revived interests in applying deep learning techniques, especially recurrent neural networks and their variants~\citep{flunkert2017deepar,wen2017multi,mukherjee2018armdn}. However, they are solely data-driven and lack a principled approach to deal with uncertainty. 

% Two general RNN forecasting models that are designed for forecasting big time series include DeepAR~\citep{flunkert2017deepar} and MQ-RNN\cite{wen2017multi}. DeepAR employs the standard RNN structure with autoregressive input and generates predictions (emission probability) with a user-specific distribution parameterized by the output of an LSTM cell. The sole source of the stochasticity in DeepAR comes from the observation model, while the model itself is deterministic, limiting the ability of the uncertainty estimation to propagate forward. MQ-RNN, on the other hand, armed with the Sequence-to-Sequence structure, directly outputs the conditional quantiles of the prediction. However, lacking a proper generative process limits its ability to generate consistent prediction (for example, a lower quantile can be larger than a higher quantile). 

Effectively combining probabilistic graphical models and deep learning approaches has been an active research area.  Several approaches have been proposed for marrying RNN with SSMs through either one or both of the following: (i) extending the Gaussian emission to complex likelihood models; (ii) making the transition equation non-linear via a multi-layer perceptron (MLP) or interlacing SSM with transition matrices temporally specified by RNNs. Deep Markov Models (DMMs), proposed by~\cite{krishnan2017structured,krishnan2015deep}, keep the Gaussian transition dynamics with mean and covariance matrix parameterized by MLPs. Stochastic RNNs (SRNNs)~\citep{fraccaro2016sequential} explicitly incorporate the deterministic dynamics from RNNs that do not depend on latent variables by interlacing them with a SSM.  \citet{chung2015recurrent} first proposed Variational RNNs (VRNNs), which is another way to make the transition equation non-linear, by cutting ties between the latent states and associating them with deterministic states.  This makes the state transition non-linearly determined by the RNN.  VRNNs are also used in Latent LSTM Allocation (LLA)~\citep{zaheer2017latent} and State-Space LSTM (SSL)~\citep{zheng2017state}. These models require expensive inference at the global level through a recognition network, with is in stark contrast with Deep Factors, where the structural assumption of the variational approximation decomposes the inference problem to local probabilistic inference that is easily parallelizable and global standard RNN learning (cf. Section \ref{sect:inference}). 

In \citet{fraccaro2017disentangled} and the recent Deep State Models~\citep{rangapuram2018}, the linear Gaussian transition structure is kept intact, so that the highly efficient Kalman filter/smoother is readily applicable. Our model differs from the former in that we avoid sampling the latent states in the ELBO, and eliminate the variance coming from the Monte-Carlo estimate of the second integral in Eqn.~\eqref{eqn:selbo}. Deep State is designed for a Gaussian likelihood with time varying SSM parameters per time series. In contrast, with time invariant local SSMs and flexible global effects, our model DF-LDS offers a parsimonious representation that can handle non-Gaussian observations. 

Deep Gaussian Processes~\citep{damaianou2013deepGP} have attracted growing interests in recent years. Inspired by GP-LVM structure~\citep{lawrence2004gaussian}, Deep GPs stacks GPs on top of latent variables, resulting in more expressive mappings. Our framework provides an alternative approach to utilize GPs more efficiently. 

Due to its flexibility of interpolating between purely local and purely global models, there are a variety of common methods that can be subsumed in our proposed model framework.  Deep Factor with one factor and no random effects, accompanied with autoregressive inputs, reduces to DeepAR~\cite{flunkert2017deepar}.  One difference in our formulation is that the scale of each time series is automatically estimated rather than pre-specified as it is in DeepAR.  Changing the emission probability to Gaussian Mixtures results in AR-MDN~\cite{mukherjee2018armdn}.  Sequence-to-Sequence models for forecasting ~\citep{wen2017multi} are another family of models that are a part of our framework.  These methods make predictions discriminatively rather than generatively.  By dropping the random effects, using GPs as the prior and removing the restriction of RNNs on the global factors, we recover the semi-parametric latent factor model~\citep{teh2005semiparametric}.  By dropping the global effects, we arrive at the standard SSM or GP.   Our newly developed general variational inference algorithm applies to both of these methods and  other normally distributed local stochastic models (cf. subsubsection \ref{subsect:nongauss}).  While we have built upon existing works, to the best of our knowledge, Deep Factors provide the first model framework that incorporate SSMs and GPs with DNNs in a systematic manner.

% The non-linear behavior is approximated by locally linear transition matrices. The so-called called \emph{Kalman Variational Auto-Encoder} (KVAE) disentangles the observations (emissions) and the latent dynamics (transitions) with VAE. By making the locally linear part outside of the standard inference routine and using a fully factorized Gaussian ``decoder,'' Kalman smoothing can be readily applied. A similar idea appeared in~\cite{johnson2016composing} where a recognition network is used to output conjugate graphical model potentials so that efficient structural inference is feasible. 

%\todo{expand the discussion and add related citations related to GPs and deep GPs.}

% Our main question is whether we can combine classical structural time
% series models with data-driven neural networks in a principled and
% efficient manner. Deep Gaussian processes (DGPs)
% \citep{damaianou2013deepGP} have shown that this may be possible. Our
% work offers an alternate combination of classical and deep learning models with
% favorable scaling properties.

\section{Experiments}
\label{sec:exp}

\begin{figure*}[t]
    \centering
    \begin{minipage}{.27\textwidth}
        \centering
        \includegraphics[width=\linewidth]{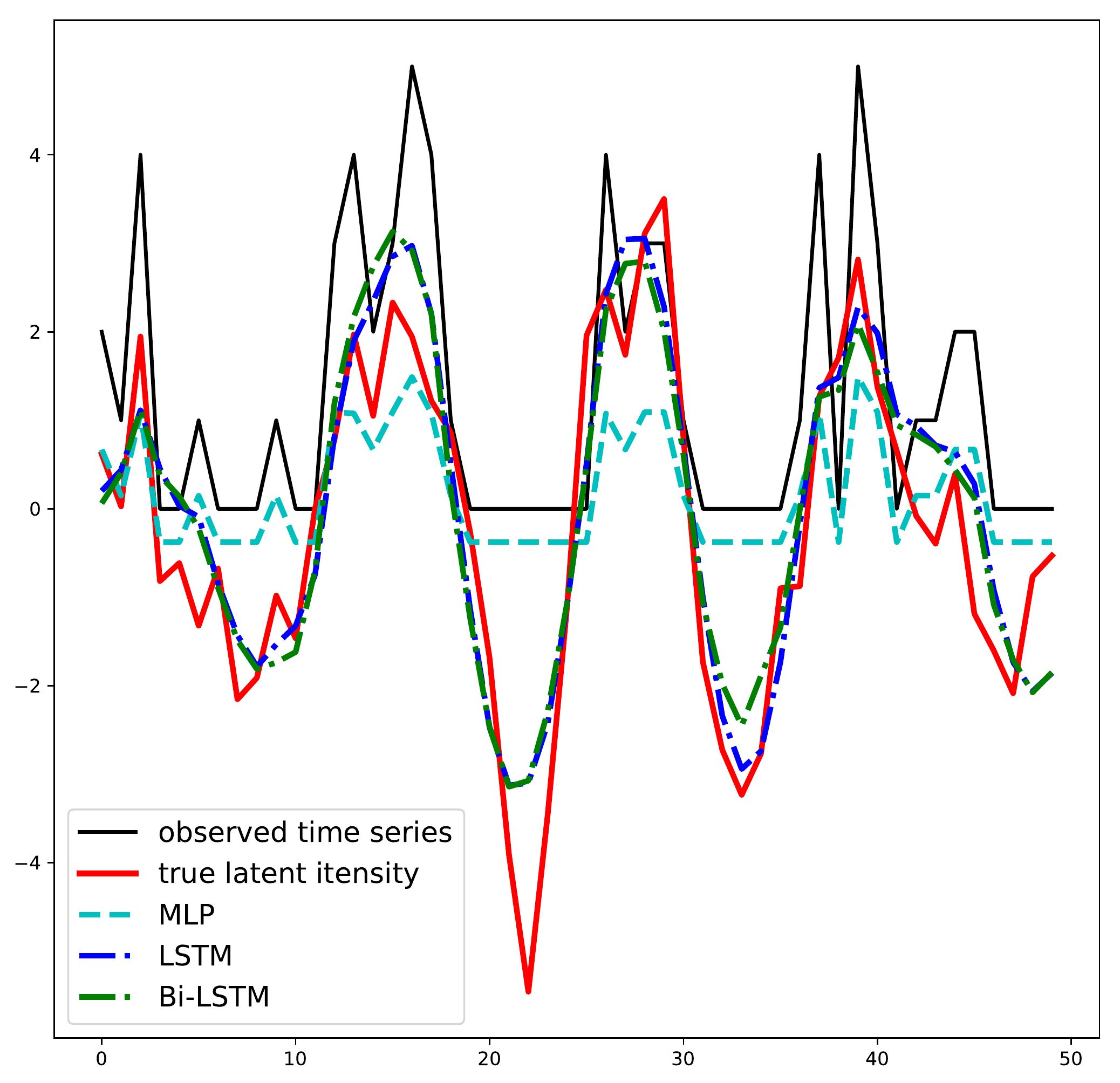} 
    \end{minipage}%
    \begin{minipage}{.27\textwidth}
        \centering
        \includegraphics[width=\linewidth]{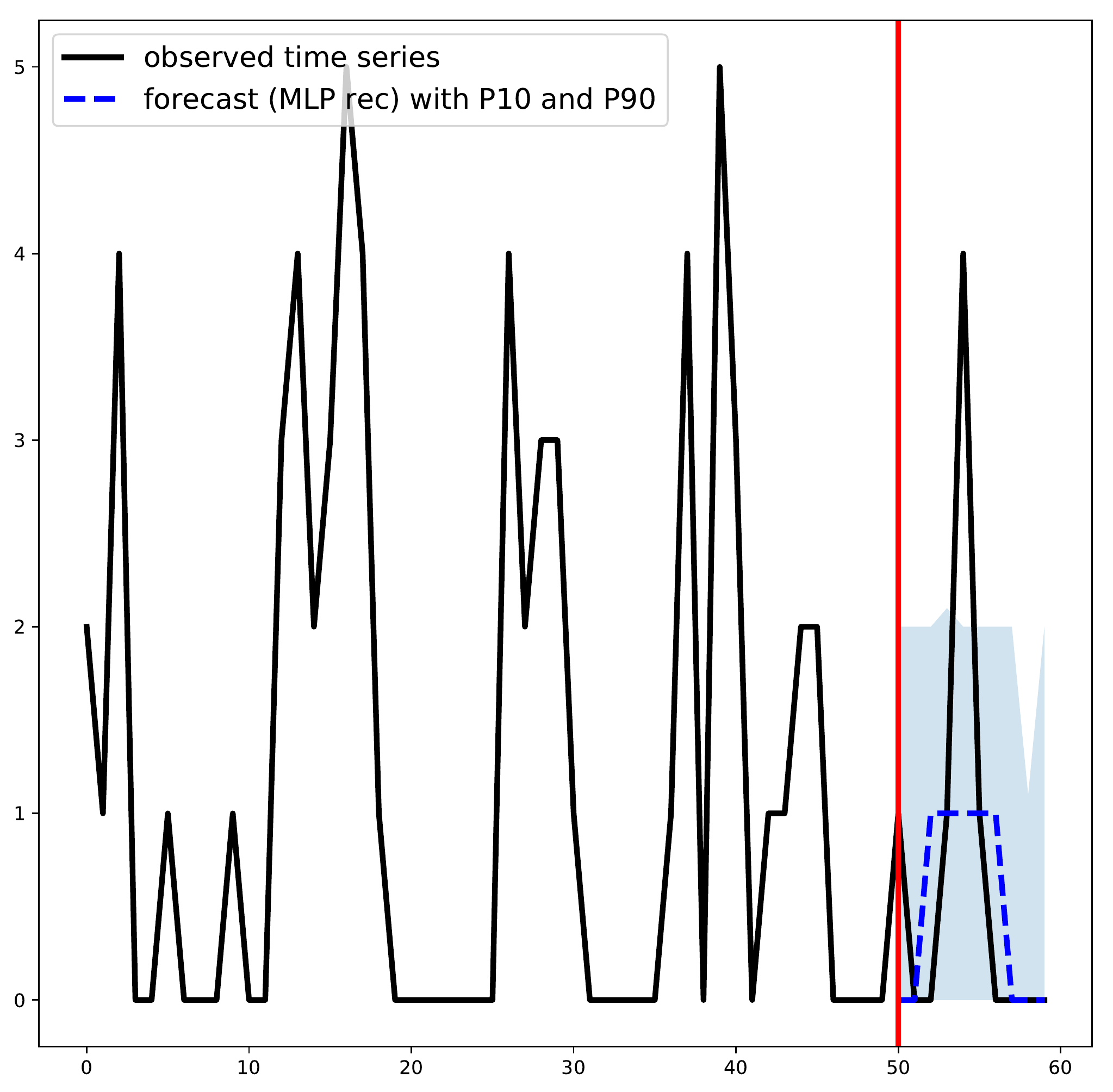}
    \end{minipage}
    \begin{minipage}{.27\textwidth}
        \centering
        \includegraphics[width=\linewidth]{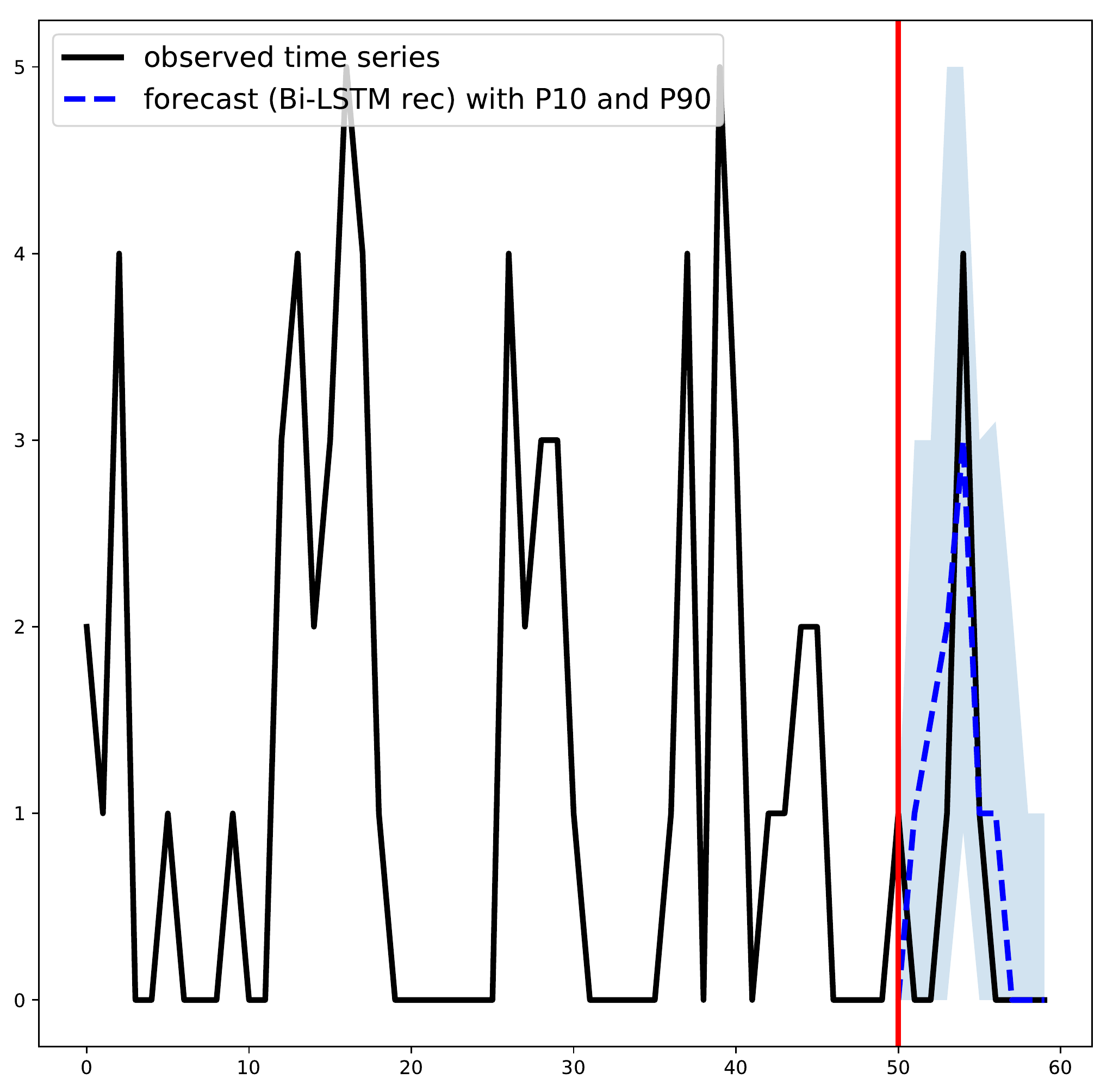} %example_u.png}                
    \end{minipage}    
    \vspace{-.15cm}
    \caption{DeepFactor (DF-LDS) with no global effects (Variational LDS). Left: reconstructed Intensity ($\ve_t$) with different recognition networks. Center and Right: predictive distributions with MLP (center) and Bidirectional LSTM (right).}
    \label{fig:recons}
    %\vspace{-.5cm}
\end{figure*}

We conduct experiments with synthetic and real-world data to provide 
evidence for the practical effectiveness of our approach. We use a p3.8xlarge 
SageMaker instance in all our experiments.  Our algorithms are implemented in MXNet Gluon~\citep{chen2015mxnet}, and make extensive use of its Linear Algebra library \cite{seeger2017auto, dai2018}. Further experiments with GPs as the local model are detailed in~\cite{maddixdeep}.
\vspace{-.2cm}
%Due to space limitations, the complete experimental section can be found in Appendix~\ref{app:empirical}. % and Appendix~
% \ref{app:synthetic} also contains
%  experiments on synthetic data that show that we can recover the underlyiing
 % global factors.
%\vspace{-.2cm}
\subsection{Model Understanding and Exploration}\label{sec:synthetic} % \textcolor{red}{
% TODOs:
% \begin{enumerate}{}
% 		\item local time series experiments (quantitatively): 1) compare different structure of recognition network; 2) shows the posterior from Stan and the recognition network for Poisson; 3) censored output experiments.
% 		\item synthetic global time series, show 1) reconstruction of the factors; 2) the forecast examples.
% 		\item point process example for taxi data (spatio-temporal).
% \end{enumerate}
% }

The first set of experiments compares our proposed structure in Eqn. ~\eqref{eqn:fixed} with no probabilistic component to the standard RNN structure on the \texttt{electricity} dataset. For each time series $i$, we have its embedding $\bm{w}_i\in\R^{10},$ and two time features $\ft_{t} \in \R^2$, day of the week and hour of the day. Given an RNN Cell (LSTM, GRU, etc.), the RNN Forecaster predicts the time series values $\hat{z}_{i,t} = \textsc{rnn}_t(\ft_{i,t}),$ where $\ft_{i,t} = [\bm{w}_i; x_t]$. Deep Factor generates the point forecast as $\hat{z}_{i,t} = \bm{w}_i^\top\textsc{rnn}_t(x_t)$. The RNN cell for RNN Forecaster has an output dimension of 1 while that of DeepFactor is of dimension 10.  The resulting number of parameters of both structures are roughly the same. 

Figure~\ref{fig:sc} demonstrates that the proposed structure significantly improves the data efficiency while having less variance.  The runtime of Deep Factor scales better than that of the standard RNN structure.  By using the concatenation of $[\bm{w}_i; x_t]$, the standard RNN structure operates on the outer product space of $\bm{x}_t$ and $\bm{w}_i$ (of dimension $12 \times T$), while Deep Factor computes the intrinsic structure (of dimension 12 and $T$ separately). 

Next, we focus on the DF-LDS model, and investigate 
(i) the effect of the recognition network for non-Gaussian 
observations with the purely local part ($f_{i,t} = 0$) (variational LDS cf. Appendix \ref{var_LDS}), and (ii) the recovery of the global 
factors in the presence of Gaussian and non-Gaussian noise. We generate data according to the following model, which is adapted from Example 24.3 in~\citep{barber2012bayesian}. 
The two-dimensional latent vector $\vh_t$ is rotated at each iteration, and then projected 
to produce a scalar observation,
$$
\vh_{t+1} = \mathbf{A}\vh_t + \bm{\epsilon}_h, \quad \mathbf{A} = \begin{bmatrix}\cos\theta & -\sin\theta\\\sin\theta & \cos\theta\end{bmatrix}, 
$$
where $\bm{\epsilon}_h \sim \mathcal{N}(0, \alpha^2\textbf{I}_2), $
$\ve_{t+1} = \bm{e}_1^T \vh_{t+1} + \epsilon_v$, $\epsilon_v \sim \mathcal{N}(0, \sigma^2)$, $\textbf{I}_2$ is the $2 \times 2$ identity matrix, and $\bm{e}_1 \in \mathbb{R}^2$ is the standard basis vector.
The true observations are generated by a Poisson distribution, 
$
\obs_t = \text{Poisson}[\lambda(\ve_t)], \ \text{where } \lambda(\ve_t) = \log[1 + \exp(\ve_t)].
$
This could be used to model seasonal products, where most of the 
sales happen when an item is in season, e.g. snowboards normally sell shortly before 
or during winters, and less so afterwards. Figure~\ref{fig:recons} shows the reconstructed intensity function $\lambda(\bm{u}_t)$, as well as corresponding forecasts for each choice of recognition network. Visual
inspections reveal that RNNs are superior over MLPs as
recognition networks. This is expected because the time series are sequential. We also test the ability of our algorithm to recover the underlying global factors. Our experiments show that even with the Poisson noise model, we are able to identify the true latent factors in the sense of distances of the subspaces spanned by the global factors (cf. Appendix~\ref{app:synthetic}).

\begin{figure}[h]
    \centering
    \begin{minipage}{.35\textwidth}
        \centering
        \includegraphics[width=\linewidth]{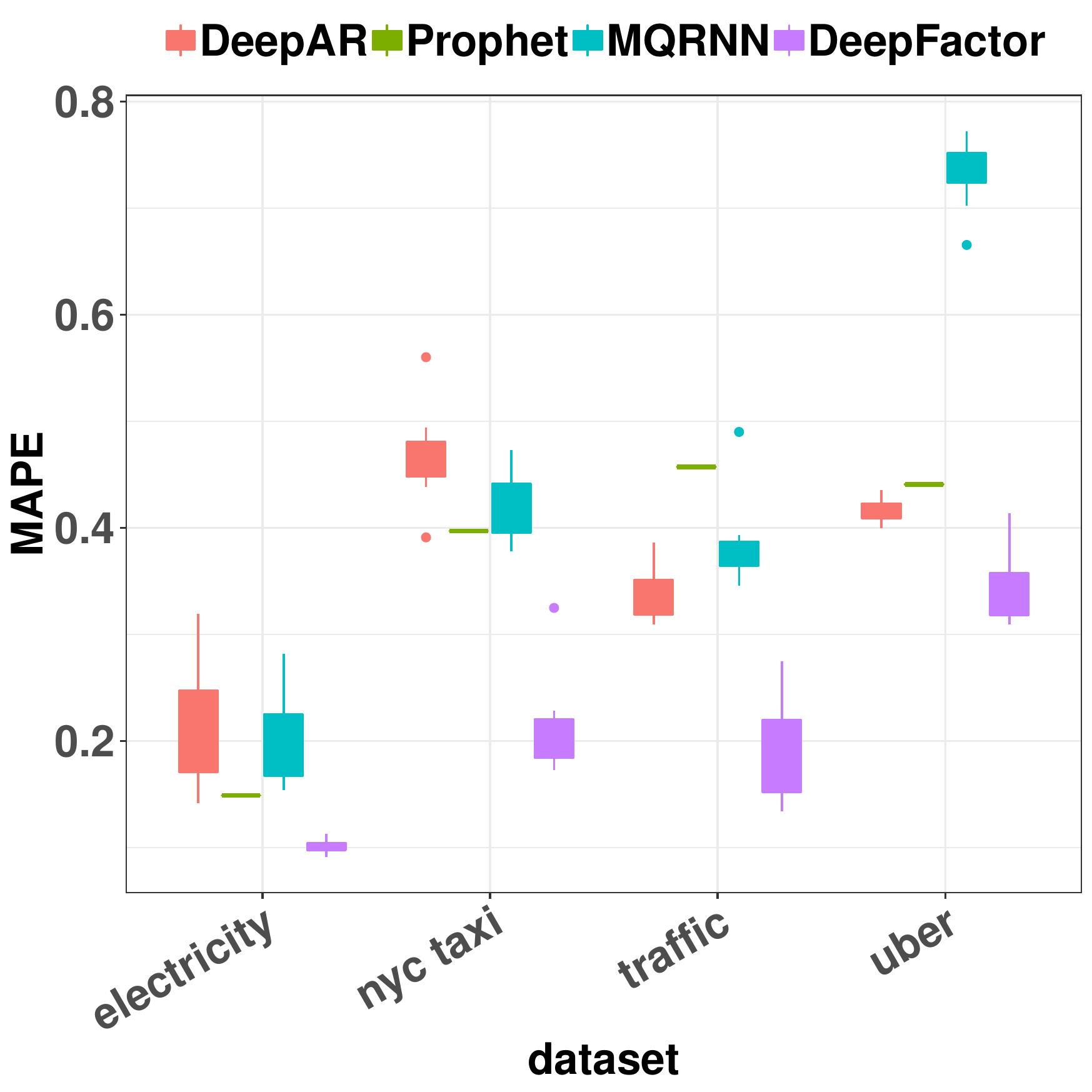} 
    \end{minipage}%
%    \begin{minipage}{.33\textwidth}
%        \centering
%        \includegraphics[width=\linewidth]{figs/p90_sub.pdf}
%    \end{minipage}
%      \begin{minipage}{.33\textwidth}
%        \centering
%        \includegraphics[width=\linewidth]{figs/RMSE_sub.pdf}
%    \end{minipage}
   % \begin{minipage}{.33\textwidth}
     %  \centering
      %  \includegraphics[width=\linewidth]{figs/RMSE_sub}
  %  \end{minipage}       
    \caption{P50QL (MAPE) results for the short-term (72-hour) forecast in Table \ref{tab:results}. Purple denotes the proposed method.}  %Prophet is a widely used statistics algorithm at Facebook, which we use to benchmark against.}
    \label{fig:selected_results}
    \vspace{-.6cm}
\end{figure}

\begin{table*}[ht]
\small
\centering
%\scriptsize
\begin{tabular}{l|c|cccc|cccc}
\toprule
\multirow{2}{*}{\textsc{ds}} & \multirow{2}{*}{\textsc{h}} & \multicolumn{4}{c}{\textsc{p50ql}} & \multicolumn{4}{c}{\textsc{p90ql}} \\
\cline{3-10}
{} & & DA & P & MR & DF & DA & P & MR & DF \\
\midrule 
\multirow{2}{*}{\textbf{E}} & 72 & 0.216\ \!$\pm$\ \!0.054 & 0.149 & 0.204\ \!$\pm$\ \!0.042 & \textbf{0.101\ \!$\pm$\ \! 0.006} &  0.182 \ \!$\pm$\ \!0.077 & 0.103 & 0.088\ \!$\pm$\ \!0.008 &\textbf{0.049}\ \!$\pm$\ \!\textbf{0.004}  \\
                            & 24 & 0.272\ \!$\pm$\ \!0.078 & 0.124 & 0.185\ \!$\pm$\ \!0.042 &\textbf{0.112\ \!$\pm$\ \!0.012} &0.100\ \!$\pm$\ \!0.013  & 0.091 & 0.083\ \!$\pm$\ \!0.008 & \textbf{0.059 \ \!$\pm$\ \!0.013}\\
\midrule
\multirow{2}{*}{\textbf{N}} & 72 & 0.468\ \!$\pm$\ \!0.043 & 0.397 & 0.418\ \!$\pm$\ \!0.031 & \textbf{0.212}\ \!$\pm$\ \!\textbf{0.044} & 0.175\ \!$\pm$\ \!0.005 & 0.178 & 0.267\ \!$\pm$\ \!0.038 &\textbf{0.104\ \!$\pm$\ \!0.028} \\
                            & 24 & 0.390\ \!$\pm$\ \!0.042 & 0.328 & 0.358\ \!$\pm$\ \!0.029 & \textbf{0.239\ \!$\pm$\ \!0.037} & 0.167\ \!$\pm$\ \! 0.005 & 0.168 & 0.248\ \!$\pm$\ \!0.088 & \textbf{0.139\ \!$\pm$\ \!0.035}\\
\midrule 
\multirow{2}{*}{\textbf{T}} & 72 & 0.337\ \!$\pm$\ \!0.026 & 0.457 & 0.383\ \!$\pm$\ \!0.040 & \textbf{0.190}\ \!$\pm$\ \!\textbf{0.048} & 0.184\ \!$\pm$\ \!0.051 & 0.207 & 0.226\ \!$\pm$\ \!0.017 & \textbf{0.127}\ \!$\pm$\ \!\textbf{0.026} \\
                            & 24 & 0.296 \ \!$\pm$\ \!0.021 & 0.380 & 0.331\ \!$\pm$\ \!0.011 & \textbf{0.225}\ \!$\pm$\ \!\textbf{0.050} & \textbf{0.149\ \!$\pm$\ \!0.011}  & 0.191 & 0.154\ \!$\pm$\ \!0.020 & 0.159\ \!$\pm$\ \!0.059 \\
\midrule
\multirow{2}{*}{\textbf{U}} & 72 & 0.417\ \!$\pm$\ \!0.011 & 0.441 & 0.730\ \!$\pm$\ \!0.031 & \textbf{0.344\ \!$\pm$\ \!0.033}  & 0.296\ \!$\pm$\ \!0.017 & 0.239 & 0.577\ \!$\pm$\ \!0.059 & \textbf{0.190\ \!$\pm$\ \!0.013}\\
                            & 24 & \textbf{0.353\ \!$\pm$\ \!0.009} & 0.468 & 0.879\ \!$\pm$\ \!0.156 & 0.425\ \!$\pm$\ \!0.063 & \textbf{0.238\ \!$\pm$\ \!0.009} & 0.239 & 0.489\ \!$\pm$\ \!0.069 & 0.238\ \!$\pm$\ \!0.026 \\
\bottomrule
\end{tabular}
\caption{Results for the short-term (72-hour) and near-term (24-hour) forecast scenarios with one week of training data.} %DA, P, MR and DF stand for DeepAR, Prophet, MQ-RNN, and DeepFactor respectively. }
\label{tab:results}
\vspace{-.3cm}
\end{table*}

\subsection{Empirical Studies} 
\label{sect-empstudies}
%break down into ucr shown in appendix
In this subsection, we test how our model performs on several real-world and publicly available datasets: \texttt{electricity} (\textbf{E}) and \texttt{traffic} (\textbf{T})
from the UCI data set~\citep{Dua:2017,yu2016temporal}, \texttt{nyc taxi} (\textbf{N})~\citep{taxi:2015} 
and \texttt{uber} (\textbf{U})~\citep{uber:2015} (cf. Appendix~\ref{app:empirical}). %These datasets are more suited for forecasting, while the additional datasets from the UCR repository ~\citep{UCRArchive} in Appendix~\ref{app:empirical} are primarily for classification purposes.  To the best of our knowledge, our experimental evaluation is one of the more comprehensive studies of the empirical performances of forecasting models. 

In the experiments, we choose the DF-RNN (\textbf{DF}) model with a Gaussian likelihood.  To assess the performance of our algorithm, we compare with DeepAR (\textbf{DA}), a state-of-art RNN-based probabilistic forecasting algorithm on the publicly available AWS SageMaker ~\citep{janu2018}, MQ-RNN (\textbf{MR}), a sequence model that generates conditional predictive quantiles ~\citep{wen2017multi}, and Prophet (\textbf{P}), a Bayesian structural time series model~\citep{taylor2017forecasting}. The Deep Factor model has 10 global factors with a LSTM cell of 1-layer and 50 hidden 
units. The noise LSTM has 1-layer and 5 hidden units. For a fair comparison with DeepAR, we use a comparable number of model parameters, that is, an embedding size of 10 with 1-layer and 50 hidden LSTM units.  The student-$t$ likelihood in DeepAR is chosen for its robust performance. The same model structure is chosen for MQ-RNN, and the decoder MLP has a single hidden layer of 20 units. We use the adam optimization method with the default parameters in Gluon to train the DF-RNN and MQ-RNN. We use the default training parameters for DeepAR. 

% For \texttt{taxi} 
% and \texttt{uber}, we use the tensor version of the DF-RNN model, TDF-RNN, with 
% log-normal and generalized Poisson model respectively and the attention networks 
% are simply the embedding for longitude and latitude.

We use the quantile loss to evaluate the 
probabilistic forecasts. For a given quantile $\rho\in(0,1)$, a target value $\obs_t$ and $\rho$-quantile prediction $\widehat{\obs}_t(\rho)$, the $\rho$-quantile 
loss is defined as
\begin{align*}
\text{QL}_\rho[\obs_t, \widehat{\obs}_t(\rho)] &= 2\big[\rho(\obs_t - \widehat{\obs}_t(\rho))\mathbb{I}_{\obs_t - \widehat{\obs}_t(\rho) > 0} \nonumber \\
&+ (1-\rho)(\widehat{\obs}_t(\rho) - \obs_t)\mathbb{I}_{\obs_t - \widehat{\obs}_t(\rho) \leqslant 0}\big].
\end{align*}
We use a normalized sum of quantile
losses, 
$\sum_{i,t} \text{QL}_\rho[z_{i,t}, \widehat{\obs}_{i,t}(\rho)] / \sum_{i,t} |z_{i,t}|,$
to compute the quantile losses for a given span across all time series.
We include results for $\rho=0.5, 0.9,$ which we abbreviate as the P50QL (mean absolute percentage error (MAPE)) and P90QL, respectively. 
%We also report the root mean square error (RMSE), which is the square  root of the aggregated squared error normalized by the product of number of time series and the length of the time series in the evaluation segment. 

For all the datasets, we limit the training length to only one week of time series (168 observations per time series).  This represents a relevant scenario that occurs frequently in demand
forecasting, where products often have only limited historical sales data. We average the Deep Factor, MQ-RNN and DeepAR results over ten trials.  We use one trial for Prophet, since classical methods are typically less variable than neural-network based models. Figure~\ref{fig:selected_results} illustrates the performances of the different algorithms in terms of the MAPE (P50 quantile loss) for the 72 hour forecast horizon.  Table~\ref{tab:results} shows the full results, and that our model outperforms the others in terms of accuracy and variability in most of the cases. For DeepAR, using SageMaker's HPO, our preliminary results (cf. Appendix~\ref{app:empirical}) show that with a larger model, it achieves a performance that is on-par with our method, which has much less parameters.  In addition, the sequence-to-sequence structure of DeepAR and MQ-RNN limits their
ability to react flexibly to changing forecasting scenarios, e.g. during 
on-demand forecasting, or interactive scenarios. For a
 forecasting scenario with a longer prediction horizon than during training horizon, DeepAR needs to be retrained to
reflect the changes in the decoder length. Similarly, they cannot generate forecasts that 
are longer than the length of the training time series, for example, the case in Figure~\ref{fig:sc}. Our method has no difficulty 
performing this task and has greater data efficiency. %, and our experiments show high accuracy even with a small number of training sequences, demonstrating efficient use of data. 

\vspace{-.2cm}
\section{Conclusion}
\vspace{-.2cm}
%\vspace{-.3cm}
We propose a novel global-local framework for forecasting a collection of related time series, accompanied with a result that uniquely characterizes exchangeable time series. Our main contribution is a
general, powerful and practically relevant modeling framework that scales, and obtains state-of-the-art performance. Future work includes comparing variational dropout~\citep{gal2016dropout} or Deep Ensemble~\cite{lakshminarayanan2017simple} of non-probabilistic DNN models (e.g., RNNForecaster (cf.~\ref{sec:synthetic})) for uncertainty. % to provide uncertainty estimation.

%It can also handle time series with different lengths while one needs to do back filling to apply MF. 

%summarize key advantages of method 

%Open question: which effect is explained by what part? Is this a problem in the optimization? 

% Another notable disadvantage of DeepAR is 
% that the model needs to be re-trained whenever the prediction length is 
% changed, while models within our framework need to be trained only once 
% and afterward can make on-demand forecast.
%\vspace{-1cm}
%Figure~\ref{fig:example2} shows several example forecast. 

% Note that all data sets we have provided here are regular in nature and therefore favor model-based approaches. Future work with other data sets will show whether our findings carry over to other, more irregular data sets.

% \begin{figure}[htp]
% \centering
%     \begin{minipage}{.48\textwidth}
%         \centering
%         \includegraphics[width=\linewidth]{figs/ex1} 
%     \end{minipage}%
%     \centering
%     \begin{minipage}{.48\textwidth}
%         \centering
%         \includegraphics[width=\linewidth]{figs/ex2}
%     \end{minipage}%
% \end{figure}

% \section{Conclusion}
% We propose a novel global-local framework for forecasting a collection of related time series and show how the models within this framework can be used to address common challenges in forecasting which include non-Gaussian data, missing data, censored output, and spatio-temporal data. Experiments on synthetic and real-world forecast dataset demonstrate the superiority of our approach. 
\newpage
\bibliography{dssm,ts}

\begin{thebibliography}{58}
\providecommand{\natexlab}[1]{#1}
\providecommand{\url}[1]{\texttt{#1}}
\expandafter\ifx\csname urlstyle\endcsname\relax
  \providecommand{\doi}[1]{doi: #1}\else
  \providecommand{\doi}{doi: \begingroup \urlstyle{rm}\Url}\fi

\bibitem[Ahmadi et~al.(2011)Ahmadi, Kersting, and Sanner]{ahmadi2011multi}
Ahmadi, B., Kersting, K., and Sanner, S.
\newblock Multi-evidence lifted message passing, with application to pagerank
  and the kalman filter.
\newblock In \emph{Twenty-Second International Joint Conference on Artificial
  Intelligence}, 2011.

\bibitem[Ahmed et~al.(2012)Ahmed, Aly, Gonzalez, Narayanamurthy, and
  Smola]{ahmed2012scalable}
Ahmed, A., Aly, M., Gonzalez, J., Narayanamurthy, S., and Smola, A.~J.
\newblock Scalable inference in latent variable models.
\newblock In \emph{Proceedings of the fifth ACM international conference on Web
  search and data mining}, pp.\  123--132. ACM, 2012.

\bibitem[Araujo et~al.(2019)Araujo, Ribeiro, Song, and
  Faloutsos]{araujo2019tensorcast}
Araujo, M., Ribeiro, P., Song, H.~A., and Faloutsos, C.
\newblock Tensorcast: forecasting and mining with coupled tensors.
\newblock \emph{Knowledge and Information Systems}, 59\penalty0 (3):\penalty0
  497--522, 2019.

\bibitem[Austin \& Panchenko(2014)Austin and Panchenko]{austin2014hierarchical}
Austin, T. and Panchenko, D.
\newblock A hierarchical version of the de finetti and aldous-hoover
  representations.
\newblock \emph{Probability Theory and Related Fields}, 159\penalty0
  (3-4):\penalty0 809--823, 2014.

\bibitem[Barber(2012)]{barber2012bayesian}
Barber, D.
\newblock \emph{Bayesian reasoning and machine learning}.
\newblock Cambridge University Press, 2012.

\bibitem[B{\"o}se et~al.(2017)B{\"o}se, Flunkert, Gasthaus, Januschowski,
  Lange, Salinas, Schelter, Seeger, and Wang]{bose2017probabilistic}
B{\"o}se, J.-H., Flunkert, V., Gasthaus, J., Januschowski, T., Lange, D.,
  Salinas, D., Schelter, S., Seeger, M., and Wang, Y.
\newblock Probabilistic demand forecasting at scale.
\newblock \emph{Proceedings of the VLDB Endowment}, 10\penalty0 (12):\penalty0
  1694--1705, 2017.

\bibitem[Brahim-Belhouari \& Bermak(2004)Brahim-Belhouari and
  Bermak]{brahim2004gaussian}
Brahim-Belhouari, S. and Bermak, A.
\newblock Gaussian process for nonstationary time series prediction.
\newblock \emph{Computational Statistics \& Data Analysis}, 47\penalty0
  (4):\penalty0 705--712, 2004.

\bibitem[Chen et~al.(2015)Chen, Li, Li, Lin, Wang, Wang, Xiao, Xu, Zhang, and
  Zhang]{chen2015mxnet}
Chen, T., Li, M., Li, Y., Lin, M., Wang, N., Wang, M., Xiao, T., Xu, B., Zhang,
  C., and Zhang, Z.
\newblock Mxnet: A flexible and efficient machine learning library for
  heterogeneous distributed systems.
\newblock \emph{arXiv preprint arXiv:1512.01274}, 2015.

\bibitem[Choi et~al.(2011)Choi, Guzman-Rivera, and Amir]{choi2011lifted}
Choi, J., Guzman-Rivera, A., and Amir, E.
\newblock Lifted relational kalman filtering.
\newblock In \emph{Twenty-Second International Joint Conference on Artificial
  Intelligence}, 2011.

\bibitem[Chung et~al.(2015)Chung, Kastner, Dinh, Goel, Courville, and
  Bengio]{chung2015recurrent}
Chung, J., Kastner, K., Dinh, L., Goel, K., Courville, A.~C., and Bengio, Y.
\newblock A recurrent latent variable model for sequential data.
\newblock In \emph{Advances in neural information processing systems}, pp.\
  2980--2988, 2015.

\bibitem[Crawley(2012)]{crawley2012mixed}
Crawley, M.~J.
\newblock Mixed-effects models.
\newblock \emph{The R Book, Second Edition}, pp.\  681--714, 2012.

\bibitem[Damianou \& N.D.(2013)Damianou and N.D.]{damaianou2013deepGP}
Damianou, A. and N.D., L.
\newblock Deep gaussian processes.
\newblock \emph{In Proceedings of the International Conference on Artificial
  Intelligence and Statistics (AISTATS)}, pp.\  207--215, 2013.

\bibitem[Dheeru \& Karra~Taniskidou(2017)Dheeru and Karra~Taniskidou]{Dua:2017}
Dheeru, D. and Karra~Taniskidou, E.
\newblock {UCI} machine learning repository.
\newblock \url{http://archive.ics.uci.edu/ml}, 2017.
\newblock University of California, Irvine, School of Information and Computer
  Sciences.

\bibitem[Diaconis(1977)]{diaconis1977_definetti}
Diaconis, P.
\newblock Finite forms of de finetti's theorem on exchangeability.
\newblock \emph{Synthese}, 36:\penalty0 271--281, 1977.

\bibitem[Diaconis \& Freedman(1980)Diaconis and
  Freedman]{diaconis1980_definetti}
Diaconis, P. and Freedman, D.
\newblock Finite exchangeable sequences.
\newblock \emph{The Annals of Probability}, 8:\penalty0 745--764, 1980.

\bibitem[Faloutsos et~al.(2018)Faloutsos, Gasthaus, Januschowski, and
  Wang]{faloutsos2018forecasting}
Faloutsos, C., Gasthaus, J., Januschowski, T., and Wang, Y.
\newblock Forecasting big time series: old and new.
\newblock \emph{Proceedings of the VLDB Endowment}, 11\penalty0 (12):\penalty0
  2102--2105, 2018.

\bibitem[Flowers(2015)]{uber:2015}
Flowers, A.
\newblock Uber {TLC} foil response.
\newblock
  \url{https://github.com/fivethirtyeight/uber-tlc-foil-response/blob/master/uber-trip-data/uber-raw-data-apr14.csv},
  2015.

\bibitem[Flunkert et~al.(2017)Flunkert, Salinas, and
  Gasthaus]{flunkert2017deepar}
Flunkert, V., Salinas, D., and Gasthaus, J.
\newblock Deepar: Probabilistic forecasting with autoregressive recurrent
  networks.
\newblock \emph{arXiv preprint arXiv:1704.04110}, 2017.

\bibitem[Forni et~al.(2000)Forni, Hallin, Lippi, and
  Reichlin]{forni2000generalized}
Forni, M., Hallin, M., Lippi, M., and Reichlin, L.
\newblock The generalized dynamic-factor model: Identification and estimation.
\newblock \emph{Review of Economics and statistics}, 82\penalty0 (4):\penalty0
  540--554, 2000.

\bibitem[Fraccaro et~al.(2016)Fraccaro, S{\o}nderby, Paquet, and
  Winther]{fraccaro2016sequential}
Fraccaro, M., S{\o}nderby, S.~K., Paquet, U., and Winther, O.
\newblock Sequential neural models with stochastic layers.
\newblock In \emph{Advances in neural information processing systems}, pp.\
  2199--2207, 2016.

\bibitem[Fraccaro et~al.(2017)Fraccaro, Kamronn, Paquet, and
  Winther]{fraccaro2017disentangled}
Fraccaro, M., Kamronn, S., Paquet, U., and Winther, O.
\newblock A disentangled recognition and nonlinear dynamics model for
  unsupervised learning.
\newblock In \emph{Advances in Neural Information Processing Systems}, pp.\
  3604--3613, 2017.

\bibitem[Gal \& Ghahramani(2016)Gal and Ghahramani]{gal2016dropout}
Gal, Y. and Ghahramani, Z.
\newblock Dropout as a bayesian approximation: Representing model uncertainty
  in deep learning.
\newblock In \emph{international conference on machine learning}, pp.\
  1050--1059, 2016.

\bibitem[Gasthaus et~al.(2019)Gasthaus, Benidis, Wang, Rangapuram, Salinas,
  Flunkert, and Januschowski]{gasthaus2019probabilistic}
Gasthaus, J., Benidis, K., Wang, Y., Rangapuram, S.~S., Salinas, D., Flunkert,
  V., and Januschowski, T.
\newblock Probabilistic forecasting with spline quantile function rnns.
\newblock In \emph{The 22nd International Conference on Artificial Intelligence
  and Statistics}, pp.\  1901--1910, 2019.

\bibitem[Gelman et~al.(2013)Gelman, Carlin, Stern, Dunson, Vehtari, and
  Rubin]{gelman2013bayesian}
Gelman, A., Carlin, J.~B., Stern, H.~S., Dunson, D.~B., Vehtari, A., and Rubin,
  D.~B.
\newblock \emph{Bayesian data analysis}.
\newblock CRC press, 2013.

\bibitem[Geweke(1977)]{geweke1977dynamic}
Geweke, J.
\newblock The dynamic factor analysis of economic time series.
\newblock \emph{Latent variables in socio-economic models}, 1977.

\bibitem[Girard et~al.(2003)Girard, Rasmussen, Candela, and
  Murray-Smith]{girard2003gaussian}
Girard, A., Rasmussen, C.~E., Candela, J.~Q., and Murray-Smith, R.
\newblock Gaussian process priors with uncertain inputs application to
  multiple-step ahead time series forecasting.
\newblock In \emph{Advances in neural information processing systems}, pp.\
  545--552, 2003.

\bibitem[Harvey(1990)]{harvey1990forecasting}
Harvey, A.~C.
\newblock \emph{Forecasting, structural time series models and the Kalman
  filter}.
\newblock Cambridge university press, 1990.

\bibitem[Hooi et~al.(2019)Hooi, Shin, Liu, and Faloutsos]{hooi2019smf}
Hooi, B., Shin, K., Liu, S., and Faloutsos, C.
\newblock Smf: Drift-aware matrix factorization with seasonal patterns.
\newblock In \emph{Proceedings of the 2019 SIAM International Conference on
  Data Mining}, pp.\  621--629. SIAM, 2019.

\bibitem[Hwang et~al.(2016)Hwang, Tong, and Choi]{hwang2016automatic}
Hwang, Y., Tong, A., and Choi, J.
\newblock Automatic construction of nonparametric relational regression models
  for multiple time series.
\newblock In \emph{International Conference on Machine Learning}, pp.\
  3030--3039, 2016.

\bibitem[Hyndman et~al.(2008)Hyndman, Koehler, Ord, and
  Snyder]{hyndman2008forecasting}
Hyndman, R., Koehler, A.~B., Ord, J.~K., and Snyder, R.~D.
\newblock \emph{Forecasting with exponential smoothing: the state space
  approach}.
\newblock Springer Science \& Business Media, 2008.

\bibitem[Januschowski et~al.(2018)Januschowski, Arpin, Salinas, Flunkert,
  Gasthaus, Stella, and Vazquez]{janu2018}
Januschowski, T., Arpin, D., Salinas, D., Flunkert, V., Gasthaus, J., Stella,
  L., and Vazquez, P.
\newblock Now available in amazon sagemaker: Deepar algorithm for more accurate
  time series forecasting.
\newblock
  \emph{https://aws.amazon.com/blogs/machine-learning/now-available-in-amazon-sagemaker-deepar-algorithm-for-more-accurate-time-series-forecasting/},
  2018.

\bibitem[Jing \& Smola(2017)Jing and Smola]{jing2017neural}
Jing, H. and Smola, A.~J.
\newblock Neural survival recommender.
\newblock In \emph{Proceedings of the Tenth ACM International Conference on Web
  Search and Data Mining}, pp.\  515--524. ACM, 2017.

\bibitem[Kingma \& Welling(2014)Kingma and Welling]{kingma2013auto}
Kingma, D.~P. and Welling, M.
\newblock Auto-encoding variational bayes.
\newblock \emph{ICLR}, 2014.

\bibitem[Krishnan et~al.(2015)Krishnan, Shalit, and Sontag]{krishnan2015deep}
Krishnan, R.~G., Shalit, U., and Sontag, D.
\newblock Deep kalman filters.
\newblock \emph{arXiv preprint arXiv:1511.05121}, 2015.

\bibitem[Krishnan et~al.(2017)Krishnan, Shalit, and
  Sontag]{krishnan2017structured}
Krishnan, R.~G., Shalit, U., and Sontag, D.
\newblock Structured inference networks for nonlinear state space models.
\newblock In \emph{AAAI}, pp.\  2101--2109, 2017.

\bibitem[Lakshminarayanan et~al.(2017)Lakshminarayanan, Pritzel, and
  Blundell]{lakshminarayanan2017simple}
Lakshminarayanan, B., Pritzel, A., and Blundell, C.
\newblock Simple and scalable predictive uncertainty estimation using deep
  ensembles.
\newblock In \emph{Advances in Neural Information Processing Systems}, pp.\
  6402--6413, 2017.

\bibitem[Lawrence(2004)]{lawrence2004gaussian}
Lawrence, N.~D.
\newblock Gaussian process latent variable models for visualisation of high
  dimensional data.
\newblock In \emph{Advances in neural information processing systems}, pp.\
  329--336, 2004.

\bibitem[Low et~al.(2011)Low, Agarwal, and Smola]{low2011multiple}
Low, Y., Agarwal, D., and Smola, A.~J.
\newblock Multiple domain user personalization.
\newblock In \emph{Proceedings of the 17th ACM SIGKDD international conference
  on Knowledge discovery and data mining}, pp.\  123--131. ACM, 2011.

\bibitem[Maddix et~al.(2018)Maddix, Wang, and Smola]{maddixdeep}
Maddix, D.~C., Wang, Y., and Smola, A.
\newblock Deep factors with gaussian processes for forecasting.
\newblock \emph{NeurIPS workshop on Bayesian Deep Learning}, 2018.

\bibitem[Mukherjee et~al.(2018)Mukherjee, Shankar, Ghosh, Tathawadekar,
  Kompalli, Sarawagi, and Chaudhury]{mukherjee2018armdn}
Mukherjee, S., Shankar, D., Ghosh, A., Tathawadekar, N., Kompalli, P.,
  Sarawagi, S., and Chaudhury, K.
\newblock Armdn: Associative and recurrent mixture density networks for eretail
  demand forecasting.
\newblock \emph{arXiv preprint arXiv:1803.03800}, 2018.

\bibitem[Pan \& Yao(2008)Pan and Yao]{pan2008modelling}
Pan, J. and Yao, Q.
\newblock Modelling multiple time series via common factors.
\newblock \emph{Biometrika}, 95\penalty0 (2):\penalty0 365--379, 2008.

\bibitem[Rangapuram et~al.(2018)Rangapuram, Seeger, Gasthaus, Stella, Wang, and
  Januschowski]{rangapuram2018}
Rangapuram, S.~S., Seeger, M., Gasthaus, J., Stella, L., Wang, Y., and
  Januschowski, T.
\newblock Deep state space models for time series forecasting.
\newblock In \emph{Advances in Neural Information Processing Systems}, 2018.

\bibitem[Rasmussen \& Williams(2006)Rasmussen and
  Williams]{rasmussen2006gaussian}
Rasmussen, C.~E. and Williams, C.~K.
\newblock \emph{Gaussian process for machine learning}.
\newblock MIT press, 2006.

\bibitem[Rezende et~al.(2014)Rezende, Mohamed, and
  Wierstra]{rezende2014stochastic}
Rezende, D.~J., Mohamed, S., and Wierstra, D.
\newblock Stochastic backpropagation and approximate inference in deep
  generative models.
\newblock In \emph{International Conference on Machine Learning}, pp.\
  1278--1286, 2014.

\bibitem[Seeger(2004)]{seeger2004gaussian}
Seeger, M.
\newblock Gaussian processes for machine learning.
\newblock \emph{International journal of neural systems}, 14\penalty0
  (02):\penalty0 69--106, 2004.

\bibitem[Seeger et~al.(2017)Seeger, Hetzel, Dai, Meissner, and
  Lawrence]{seeger2017auto}
Seeger, M., Hetzel, A., Dai, Z., Meissner, E., and Lawrence, N.~D.
\newblock Auto-differentiating linear algebra.
\newblock \emph{arXiv preprint arXiv:1710.08717}, 2017.

\bibitem[Seeger et~al.(2016)Seeger, Salinas, and Flunkert]{seeger2016bayesian}
Seeger, M.~W., Salinas, D., and Flunkert, V.
\newblock Bayesian intermittent demand forecasting for large inventories.
\newblock In \emph{Advances in Neural Information Processing Systems}, pp.\
  4646--4654, 2016.

\bibitem[Stock \& Watson(2011)Stock and Watson]{stock2011dynamic}
Stock, J.~H. and Watson, M.
\newblock Dynamic factor models.
\newblock \emph{Oxford handbook on economic forecasting}, 2011.

\bibitem[Taxi \& Commission(2015)Taxi and Commission]{taxi:2015}
Taxi, N. and Commission, L.
\newblock {TLC} trip record data.
\newblock \url{http://www.nyc.gov/html/tlc/html/about/trip_record_data.shtml},
  2015.

\bibitem[Taylor \& Letham(2017)Taylor and Letham]{taylor2017forecasting}
Taylor, S.~J. and Letham, B.
\newblock Forecasting at scale.
\newblock \emph{The American Statistician}, 2017.

\bibitem[Teh et~al.(2005)Teh, Seeger, and Jordan]{teh2005semiparametric}
Teh, Y., Seeger, M., and Jordan, M.
\newblock Semiparametric latent factor models.
\newblock In \emph{AISTATS 2005-Proceedings of the 10th International Workshop
  on Artificial Intelligence and Statistics}, pp.\  333--340, 2005.

\bibitem[Wen et~al.(2017)Wen, Torkkola, and Narayanaswamy]{wen2017multi}
Wen, R., Torkkola, K., and Narayanaswamy, B.
\newblock A multi-horizon quantile recurrent forecaster.
\newblock \emph{NIPS Workshop on Time Series}, 2017.

\bibitem[Xie et~al.(2017)Xie, Tank, Greaves-Tunnell, and Fox]{xie2017unified}
Xie, C., Tank, A., Greaves-Tunnell, A., and Fox, E.
\newblock A unified framework for long range and cold start forecasting of
  seasonal profiles in time series.
\newblock \emph{arXiv preprint arXiv:1710.08473}, 2017.

\bibitem[Xu et~al.(2009)Xu, Kersting, and Tresp]{xu2009multi}
Xu, Z., Kersting, K., and Tresp, V.
\newblock Multi-relational learning with gaussian processes.
\newblock In \emph{Twenty-First International Joint Conference on Artificial
  Intelligence}, 2009.

\bibitem[Yu et~al.(2016)Yu, Rao, and Dhillon]{yu2016temporal}
Yu, H.-F., Rao, N., and Dhillon, I.~S.
\newblock Temporal regularized matrix factorization for high-dimensional time
  series prediction.
\newblock In \emph{Advances in neural information processing systems}, pp.\
  847--855, 2016.

\bibitem[Zaheer et~al.(2017)Zaheer, Ahmed, and Smola]{zaheer2017latent}
Zaheer, M., Ahmed, A., and Smola, A.~J.
\newblock Latent lstm allocation: Joint clustering and non-linear dynamic
  modeling of sequence data.
\newblock In \emph{International Conference on Machine Learning}, pp.\
  3967--3976, 2017.

\bibitem[Zheng et~al.(2017)Zheng, Zaheer, Ahmed, Wang, Xing, and
  Smola]{zheng2017state}
Zheng, X., Zaheer, M., Ahmed, A., Wang, Y., Xing, E.~P., and Smola, A.~J.
\newblock State space lstm models with particle mcmc inference.
\newblock \emph{arXiv preprint arXiv:1711.11179}, 2017.

\bibitem[Zhenwen et~al.(2018)Zhenwen, Meissner, and Lawrence]{dai2018}
Zhenwen, D., Meissner, E., and Lawrence, N.~D.
\newblock Mxfusion: A modular deep probabilistic programming library.
\newblock \emph{NIPS 2018 Workshop MLOSS}, 2018.

\end{thebibliography}
\bibliographystyle{icml2019}

\section*{Acknowledgments}
We would like to thank the reviews, Valentin Flunkert, David Salinas, Michael Bohlke-schneider, Edo Liberty and Bing Xiang for their feedbacks.

\onecolumn
\appendix
%!TEX root = factor_icml.tex
\counterwithin{figure}{section}
\counterwithin{table}{section}

\section{Deep Factor Models with Random Effects}
\label{seq:dfm}
In this section, we provide additional details on the models discussed in Section~\ref{sect:genmodel}. In what follows, we overload the notation and use $\bm{g}_t(\cdot): \R^d \rightarrow \R^K$ to denote the vector function consisting of the global factors.
\begin{table*}[ht]
\centering
\begin{tabular}{l|m{4cm}|c|ccc}
\toprule
\textsc{name} &  \textsc{description} & \textsc{local} & \textsc{likelihood (gaussian case)}\\
\midrule
DF-RNN & Zero-mean Gaussian noise process given by \textsc{rnn} & $r_{i,t} \sim \mathcal{N}(0, \sigma_{i,t}^2)$ & $p(\obs_i) = \prod_{t}\mathcal{N}(z_{i,t} - f_{i,t} | 0, \sigma_{i,t}^2)$\\
\midrule
DF-LDS & State-space models & $r_{i,t} \sim \text{LDS}_{i,t}$ (cf. Eqn. \eqref{eqn:df_issm}) & $p(\obs_i)$ given by Kalman Filter\\
\midrule
DF-GP & Zero-mean Gaussian Process  & $r_{i,t}\sim \text{GP}_{i,t}$ (cf. Eqn. \eqref{eqn:df_gp})& $p(\obs_i) = \mathcal{N}(\obs_i -  \bm{f}_i| \bm{0}, \mathbf{K}_{i} + \bm{\sigma}_i^2\textbf{I})$\\
\bottomrule
\end{tabular}
%\caption{Summary of the datasets used in this paper. Both DF-RNN and TDF-RNN are described in Section~\ref{sec:df-rnn}.}
\caption{Summary table of Deep Factor Models with Random Effects. The likelihood column is under the assumption of Gaussian noise. }%The notation $\ominus$ means either subtraction or division.}
\end{table*}

\subsection{DF-RNN: Gaussian noise process as the local model} 
\label{sec:df-rnn}
The local model in DF-RNN is zero-mean i.i.d. Gaussian, where the variance $\sigma_{i,t}^2$ is generated by a noise RNN process. The generative model is given as: 
\begin{equation*}
\begin{split}
\text{random effect}:\quad & \varepsilon_{i,t} \sim \mathcal{N}(0, \sigma_{i,t}^2), \quad \sigma_{i,t} := \textsc{rnn}(x_{i,t}),\\
\text{fixed effect}:\quad  &f_{i,t} = \bm{w}_i^\top \bm{g}_t(x_{i,t}),\\
\text{emission}: \quad & z_{i,t} \sim p(\cdot|u_{i,t}),\ u_{i,t} = f_{i,t} + r_{i,t}.
\label{eqn:df_rnn}
\end{split}
\end{equation*}
% \begin{enumerate}
% \item Generate the noise standard deviation sequence $\sigma_{1:T}$ with $\textsc{rnn}_\epsilon$; 
% \item Sample the random effects, 
% \[
% \varepsilon_t \sim \mathcal{N}(0, \sigma_t), 
% \]
% \item Sample the observations such that
% \[
% z_t \sim p_\theta(u_t), \quad u_t = w^\top g_t + \varepsilon_t,\quad, t=1, \ldots, T.
% \]
% \end{enumerate}
Although the noise is i.i.d., there is correlated noise from the latent RNN process. For the non-Gaussian likelihood, the latent innovation terms $\bm{\varepsilon}$ in the latent function $\ve$ requires inference.  To do so, we use Algorithm~\ref{alg:1}. The variational lower bound in Eqn. \eqref{eqn:selbo} is straightforward to calculate, since the marginal likelihood $\log p(\widetilde{\ve})$ is the log-likelihood of a multivariate Gaussian with mean $w^\top g_t$ and a covariance matrix with $\varepsilon_t^2$ on the diagonal entries. DF-RNN can be seen as a type of deep Gaussian latent variable model. 

\subsection{DF-LDS: LDS as the local model}
\label{sec:df-lds}
The generative model for DF-LDS is given as:
\begin{equation*}
\begin{split}
\text{random effect}:\quad & \vh_{i,t} = \mathbf{F}_{i,t} \vh_{i,t-1} + \vq_{i,t} \epsilon_{i,t}, \quad \epsilon_{i,t} \sim \mathcal{N}(0, 1), \\
& r_{i,t} = \va_{i,t}^\top \vh_{i,t} \\ %+ \nu_{i,t} \\ %, \quad \epsilon_{i,t} \sim \mathcal{N}(0, 1)  \\
%& \\%\quad \nu_{i,t} \sim \mathcal{N}(0, \sigma^2_{i,t})\\
\text{fixed effect}:\quad  & f_{i,t} = \bm{w}_i^\top \bm{g}_t(x_{i,t}),\\
\text{emission}: \quad & z_{i,t} \sim p(\cdot|u_{i,t}),\ u_{i,t} = f_{i,t} + r_{i,t}.
\label{eqn:df_issm2}
\end{split}
\end{equation*}

We consider the level-trend ISSM model with damping,
\[
\va_{i,t} = \begin{bmatrix} \delta_i \\ \gamma_i\end{bmatrix}, \quad \mathbf{F}_{i,t} = \begin{bmatrix} \delta_i & \gamma_i \\ 0 & \gamma_i \end{bmatrix},\quad \text{and}\quad \vq_{i,t} = \begin{bmatrix} \alpha_{i} \\ \beta_{i} \end{bmatrix}, 
\]
where $\delta_i, \gamma_i$ control the damping effect, and the parameters $\alpha_{i} > 0$ and $\beta_{i} > 0$ contain the innovation strength for the level and trend (slope), respectively. The initial latent state $h_{i,0}$ is assumed to follow an isotropic Gaussian distribution. We choose a simple level-trend ISSM because we expect the global factors with attention to explain the majority of the time series' dynamics, such as trend and seasonality. This avoids having to hand design the latent states of the ISSM. 

As a special case, we recover SSM with non-Gaussian likelihood, and our inference algorithm gives a new approach to perform inference on such models, which we call \emph{Variational LDS. }

\paragraph{Variational LDS: DF-LDS with no global factors}
\label{var_LDS}
Variational LDS is a state-space model (SSM) with an arbitrary likelihood as its emission function. In this subsection, we supply additional details of the inference of the variational LDS using the synthetic example in Section~\ref{sec:synthetic}.  The data is generated as,
$$
h_{t+1} = Ah_{t} + \epsilon_h, \quad A = \begin{bmatrix}\cos\theta & -\sin\theta\\\sin\theta & \cos\theta\end{bmatrix}, 
$$
$$
\epsilon_h \sim \mathcal{N}(0, \alpha^2\textbf{I}_2), \quad v_{t+1} = e_1^Th_{t+1} + \epsilon_v, \quad \epsilon_v \sim \mathcal{N}(0, \sigma^2).
$$
Our observations are generated as:
$$
z_{t} = \text{Poisson}[\lambda(v_t)], \qquad \lambda(v_t) = \log[1 + \exp(v_t)], 
$$
resulting in the popular Poisson LDS. 

Our goal is perform inference over the latent variables, $\{h, \ve\}$, where $h = h_{1:T}$ and $\ve = v_{1:T}$, given the observations $z := z_{1:T}$. As in Section \ref{subsect:nongauss}, we use a variational auto-encoder (VAE) with structural approximation, and a variational posterior to approximate the true posterior $p(h, \ve|z)$ as
$
q_\phi(h, \ve|z) = q_\phi(\ve|z)p(h|\ve).
$
As in DF-LDS, the first term is generated by the recognition network, and the second term chosen to match the conditional posterior. For the training procedure, we resort to Algorithm~\ref{alg:1} with no global factors, i.e., $\widetilde{r}$ is the sample generated by the recognition network. 

There are different neural networks structures to choose from for the variational encoder $q_\phi(h|\obs)$.  We choose the bi-directional LSTM since it considers information from both the past and the future.  This is similar to the backwards message in the Kalman smoother. Similar preference can also be found in~\citep{krishnan2015deep}, although their model structure differs. The recognition network is only used in the training phase to better approximate the posterior distribution of the latent function values, and so there is no information leak for the desired forecasting case. 

\subsection{DF-GP: Gaussian Process as the local model}
With DF-GP, the random effects is assumed to be drawn from a zero-mean Gaussian process. 
\begin{equation*}
\begin{split}
\text{random effect}:\quad & r_{i}(\cdot) \sim \text{GP}(0, \mathcal{K}_i(\cdot, \cdot)), \\
\text{fixed effect}:\quad  & f_{i,t} = \bm{w}_i^\top \bm{g}_t(x_{i,t}),\\
\text{latent function values}: \quad & u_{i,t} = f_{i,t} + r_{i,t}, \quad r_{i,t} = r_i(\ft_{i,t}), \\
\text{emission}: \quad & z_{i,t} \sim p(\cdot|u_{i,t})).
\label{eqn:df_issm2}
\end{split}
\end{equation*}
Simiarily with Variational LDS, with the proposed inference algorithm, we get a new approach for doing approximate inference for GP with non-Gaussian likelihood. We leave the comparison with classical approaches~\citep{rasmussen2006gaussian} such as Laplace approximation, Expectation Propagation, Variational Inference to future work.

\section{Detailed Experimental Analysis}
In this appendix, we provide further results in both the synthetic and empirical cases.
\subsection{Synthetic Experiments}
The local model and observations are generated using the variational-LDS example in Appendix \ref{var_LDS}.  We want to test the ability of the algorithm to recover the underlying latent factors. To do so, we 
add global factors that are generated from Fourier series of different orders, to the synthetic dataset described in Section~\ref{sec:synthetic}.  For
each time series, the coefficients of the linear combination are sampled 
from a uniform distribution. 

\begin{figure}[H]
    \centering
    \begin{minipage}{.28\textwidth}
        \centering
        \includegraphics[width=\linewidth]{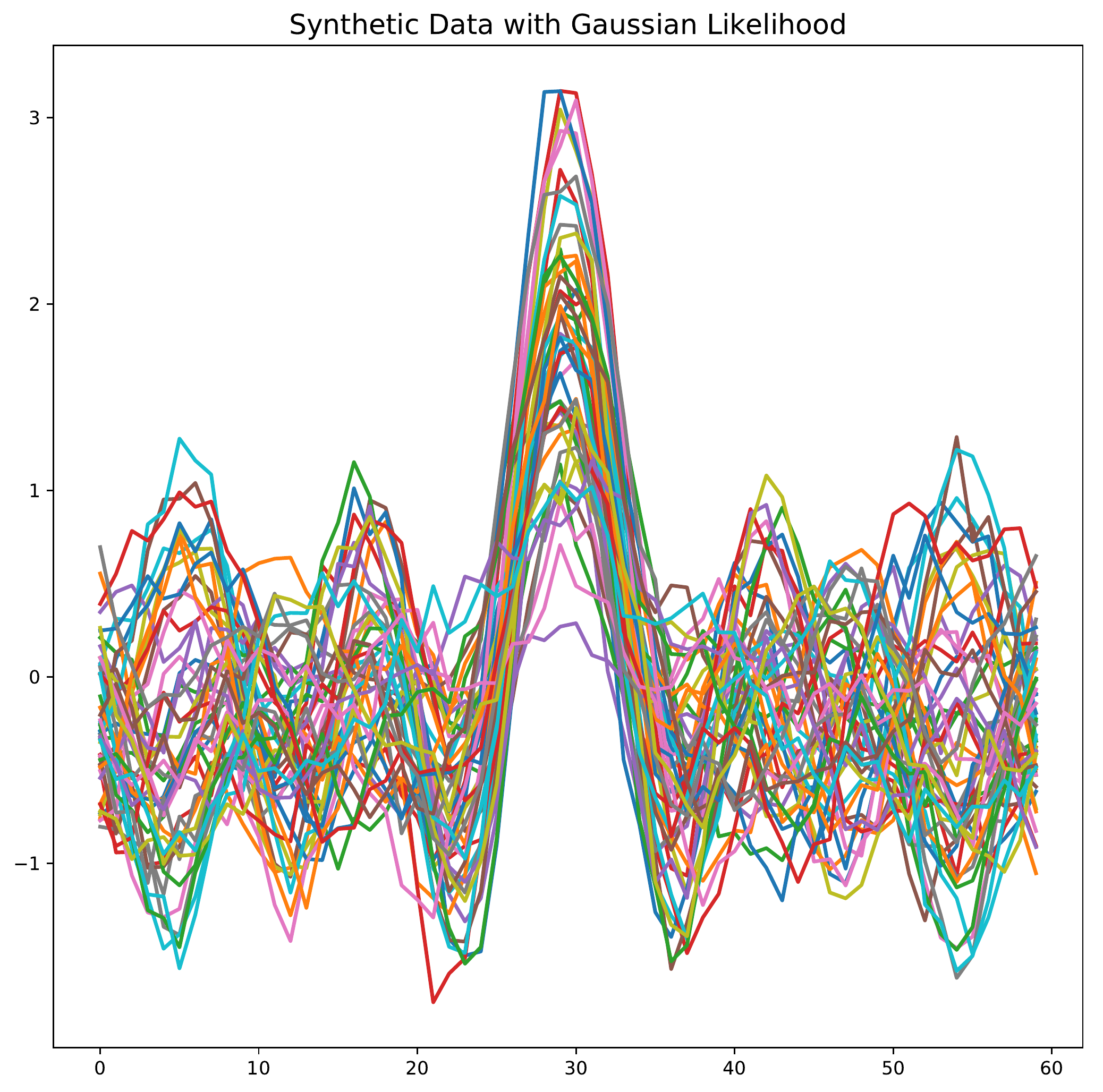} 
    \end{minipage}%
    \begin{minipage}{.28\textwidth}
        \centering
        \includegraphics[width=\linewidth]{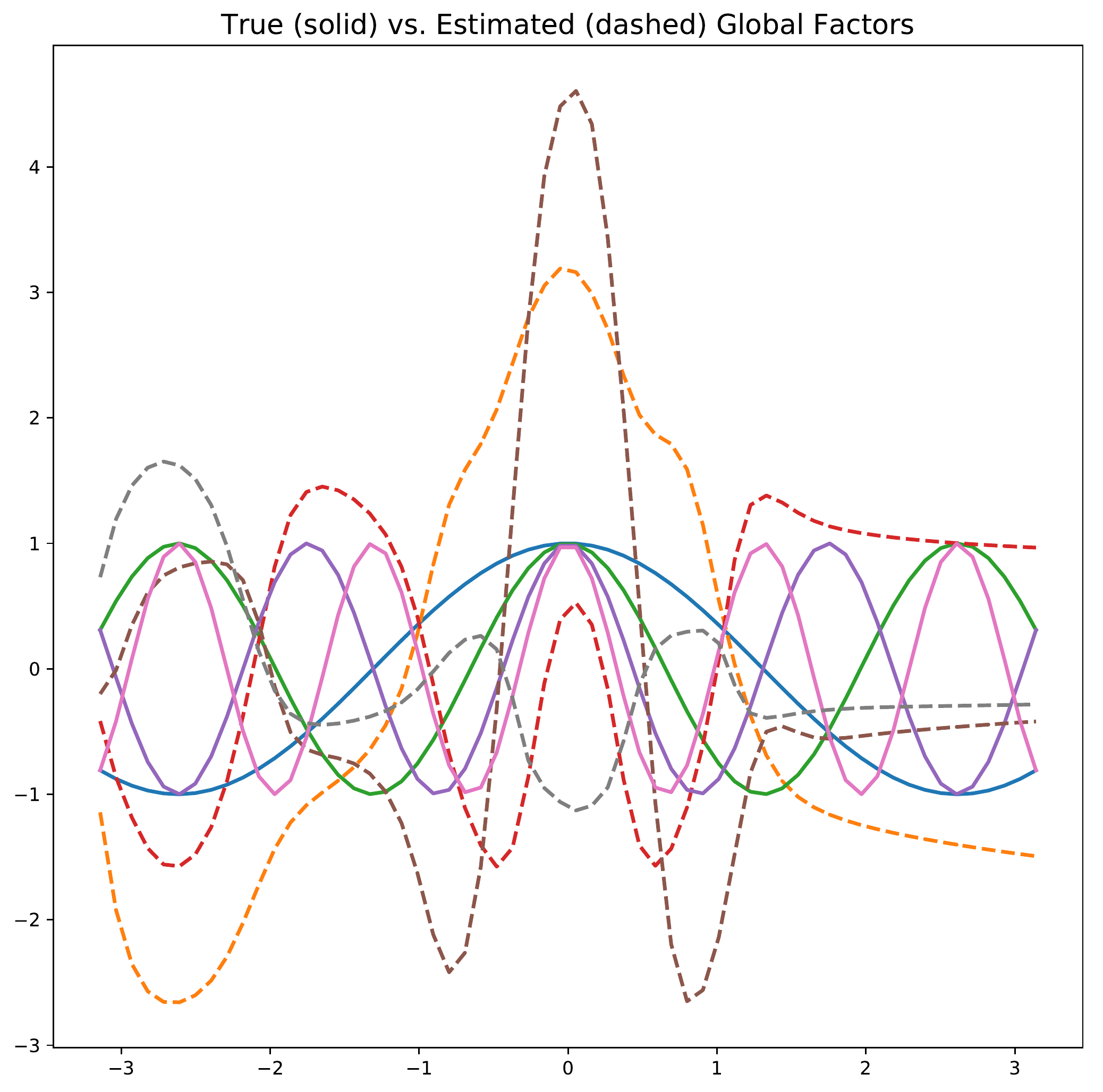}
    \end{minipage}   
    \caption{Synthetic data with Gaussian noise and true (solid) vs. estimated (dashed) global factors.}
    \label{fig:recons_factor_gauss}
    %\vspace{-.5cm}
\end{figure}

\begin{figure}[H]
    \centering
    \begin{minipage}{.28\textwidth}
        \centering
        \includegraphics[width=\linewidth]{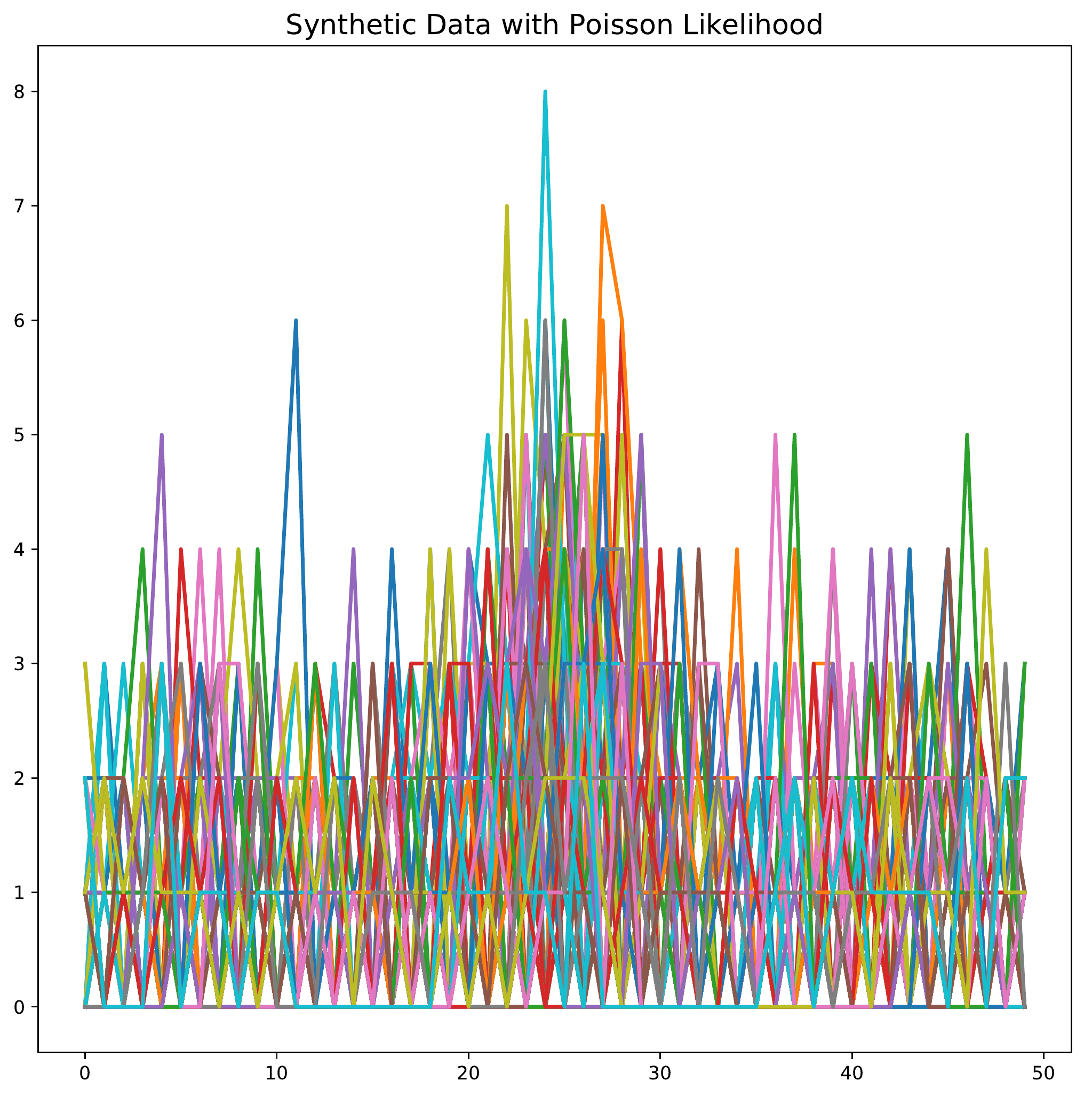}                
    \end{minipage} 
    \begin{minipage}{.28\textwidth}
        \centering
        \includegraphics[width=\linewidth]{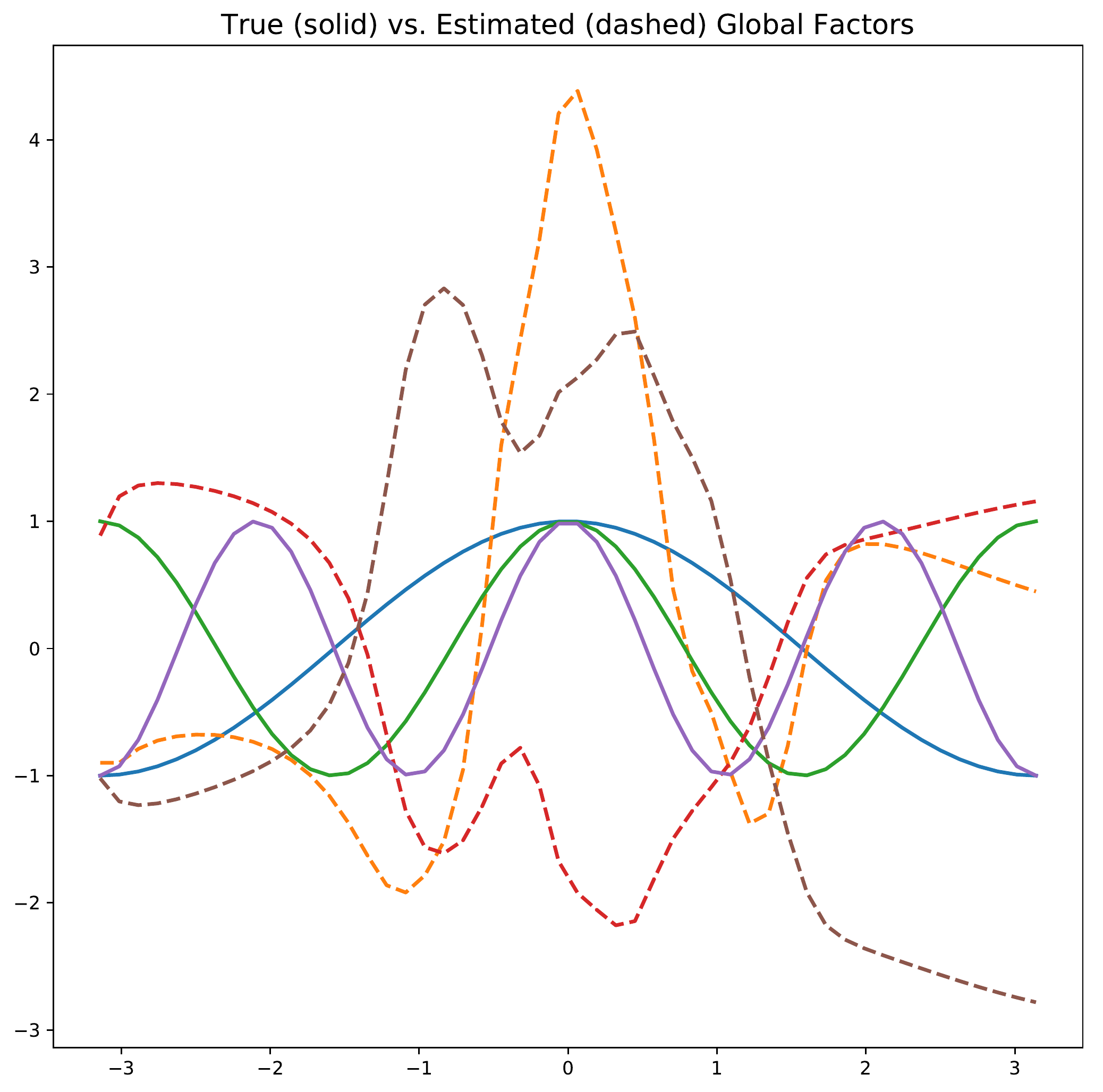}
    \end{minipage}       
    \caption{Synthetic data with Poisson noise and true (solid) vs. estimated (dashed) global factors.}
    \label{fig:recons_factor_poisson}
    %\vspace{-.5cm}
\end{figure}

 In the presence of Gaussian LDS noise 
($\ve_t$ observed), Figure~\ref{fig:recons_factor_gauss} 
shows that the algorithm does not precisely recover the global factors. The distance between the subspaces spanned by the true 
and estimated factors reveals that they are reasonably close.  Even with Poisson 
observations in Figure~\ref{fig:recons_factor_poisson}, 
the method is still able to roughly recover the factors. 

\label{app:synthetic}

\label{app:synthetic}

\subsection{Empirical Studies}
\setcounter{table}{0}
\label{app:empirical}

Table~\ref{tab:ds} summarizes the datasets that are used to test our Deep Factor models.

\begin{table}[H]
\small
\centering
\begin{tabular}{l|cccm{6cm}}
\toprule
\textsc{name} & \textsc{support} & \textsc{granularity} & \textsc{no. ts}  & \textsc{comment}\\
\midrule
\texttt{electricity} & $\R^+$ & hourly & 370 & electricity consumption of different customers \\
\texttt{traffic} & $[0, 1]$ & hourly & 963 & occupancy rate of SF bay area freeways \\
%\texttt{parts} & $\N^+$ & monthly & 1046 & auto-parts manufacture \\
\texttt{taxi} & $\N^+$ & hourly & 140 & No. of taxi pick ups in different blocks of NYC \\ 
\texttt{uber} & $\N^+$ & hourly & 114 & No. of Uber pick ups in different blocks of NYC \\ 
\bottomrule
\end{tabular}
\caption{Summary of the datasets used to test the models.}
\label{tab:ds}
\end{table}

We first compare our two methods, DF-RNN and DF-LDS with level-trend ISSM, on the \texttt{electricity} dataset with forecast horizon 24.  Table \ref{table:method_comp} shows the averaged results over 50 trials. We see that both methods have similar accuracy with DF-RNN being slightly preferrable.  Due to its simplicity and computational advantages, we choose DF-RNN as the Deep Factor model to compare against the other baseline methods. Regarding DF-GP, although it shows a good performance in our perliminary results, still extra exploration is needed for which we left to future work. 
\begin{table}[H]
\small
\centering
\begin{tabular}{c|r|r|r}
\toprule
\textsc{method} & \textsc{P50QL} & \textsc{P90QL} & \textsc{RMSE}\\
\midrule
DF-RNN & 0.144 $\pm$ 0.019 & 0.101 $\pm$ 0.044 & 888.285 $\pm$ 278.798\\
\midrule 
DF-LDS & 0.153 $\pm$ 0.016 & 0.135 $\pm$ 0.052 & 1012.551 $\pm$ 240.118\\ 
\bottomrule
\end{tabular}
\caption{Comparison of DF-RNN and DF-LDS with level-trend ISSM on \texttt{electricity} over 50 runs.}
\label{table:method_comp}
\end{table} 

\begin{figure}[h]
\centering
    \begin{minipage}{.24\textwidth}
        \centering
        \includegraphics[width=\linewidth]{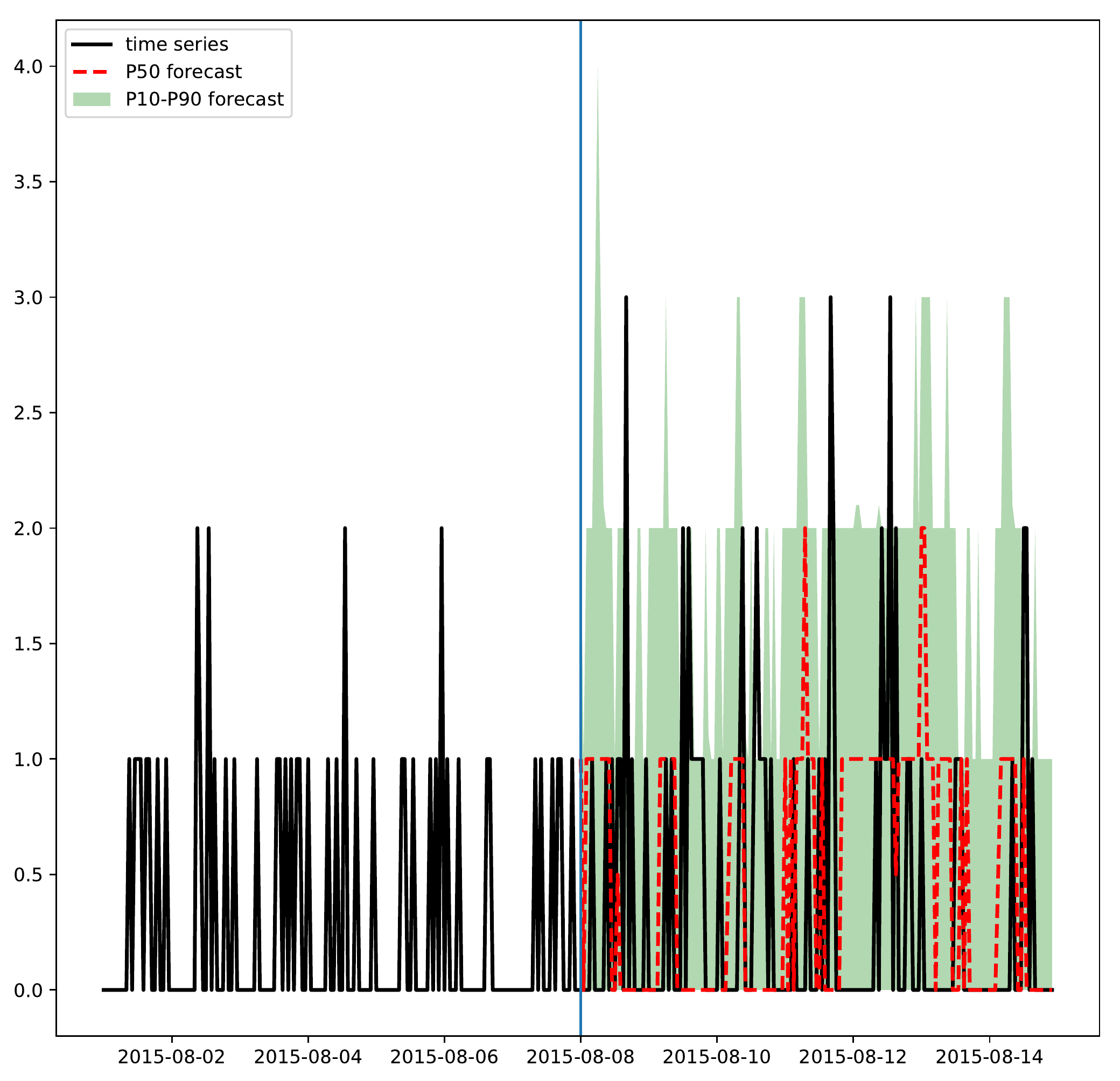} 
    \end{minipage}%
    \centering
    \begin{minipage}{.24\textwidth}
        \centering
        \includegraphics[width=\linewidth]{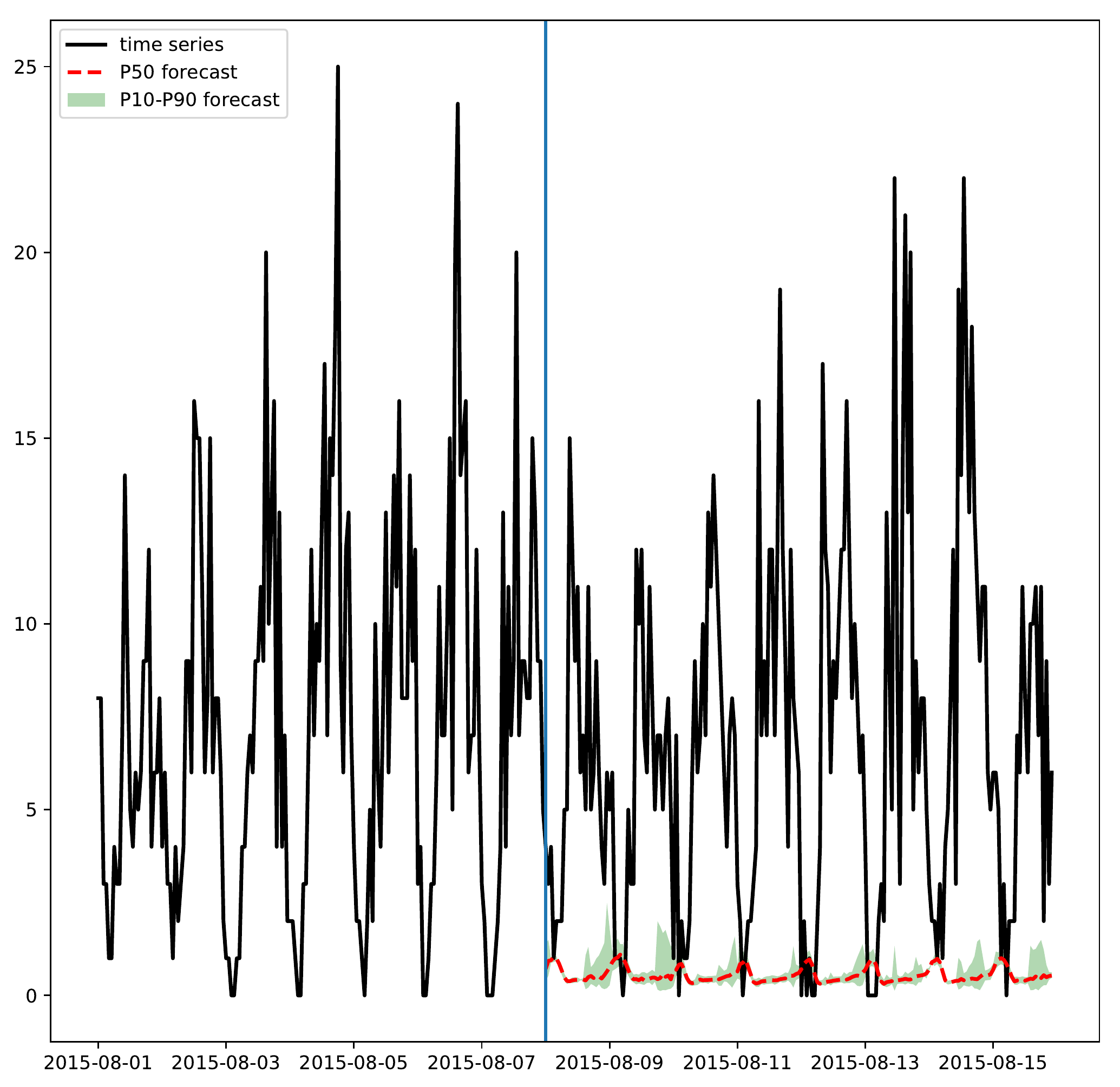} %example_u.png}                
    \end{minipage}%
    \begin{minipage}{.24\textwidth}
        \centering
        \includegraphics[width=\linewidth]{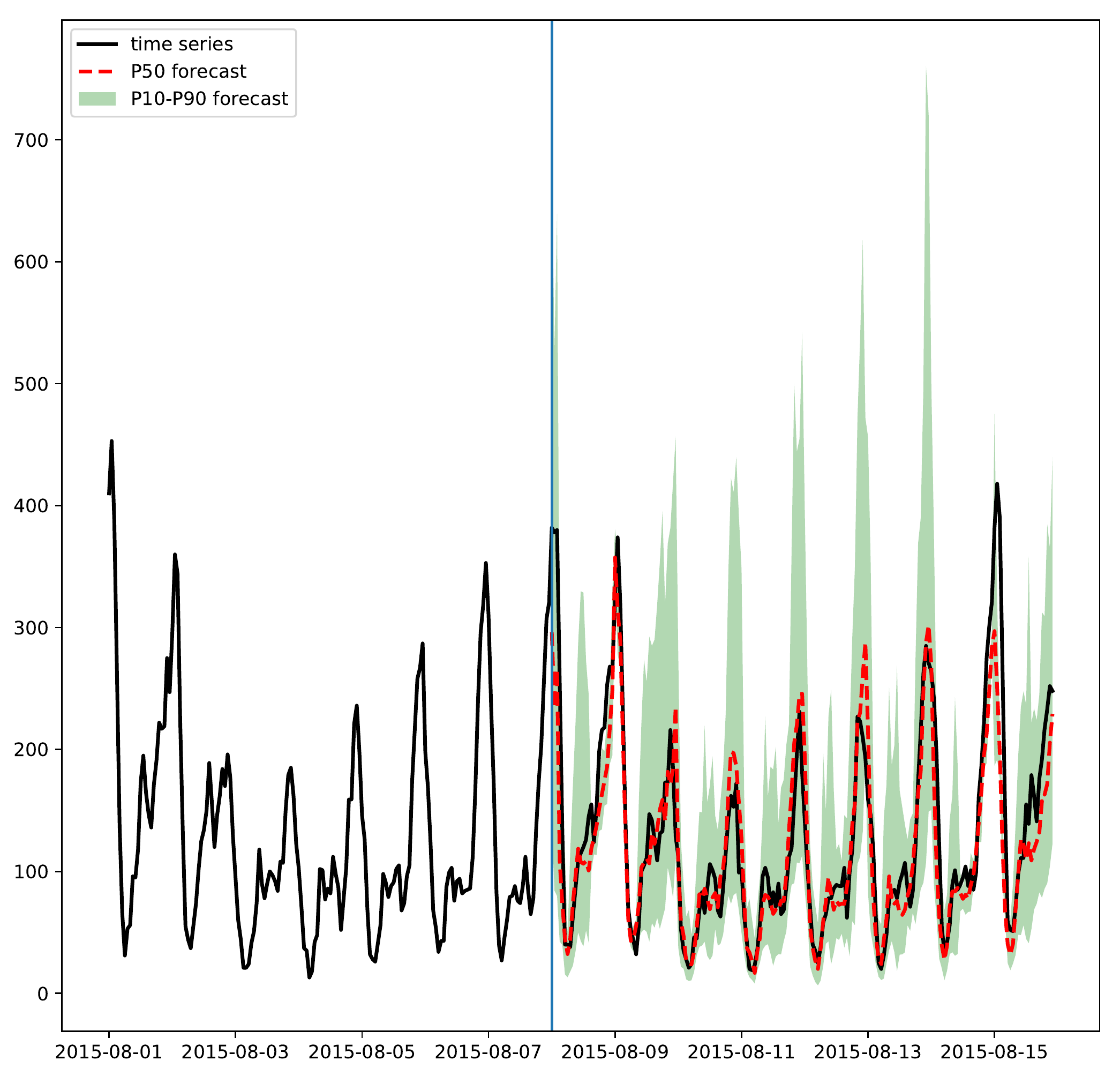}
    \end{minipage}
    \begin{minipage}{.24\textwidth}
        \centering
        \includegraphics[width=\linewidth]{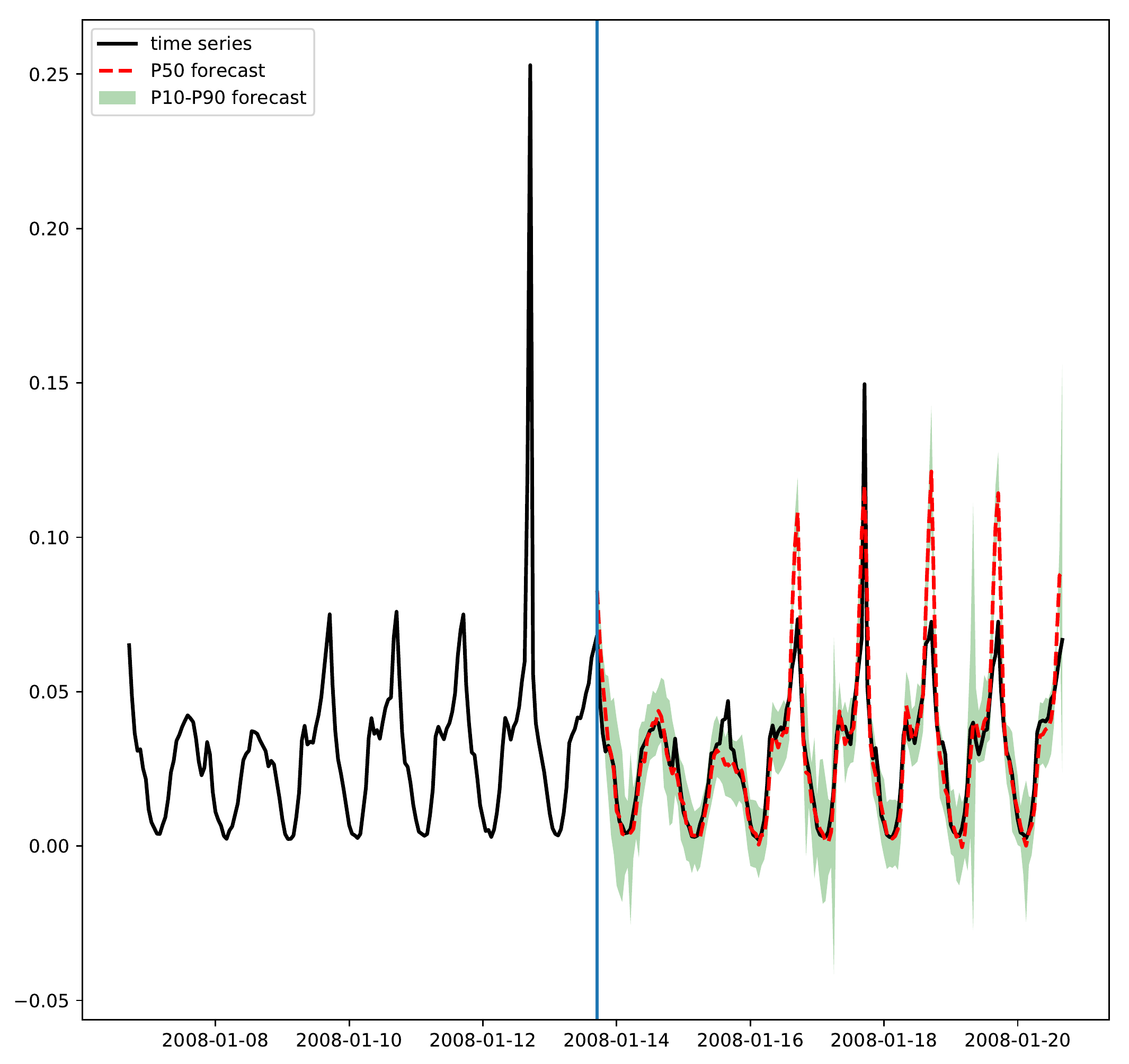}
    \end{minipage}
    \begin{minipage}{.24\textwidth}
        \centering
        \includegraphics[width=\linewidth]{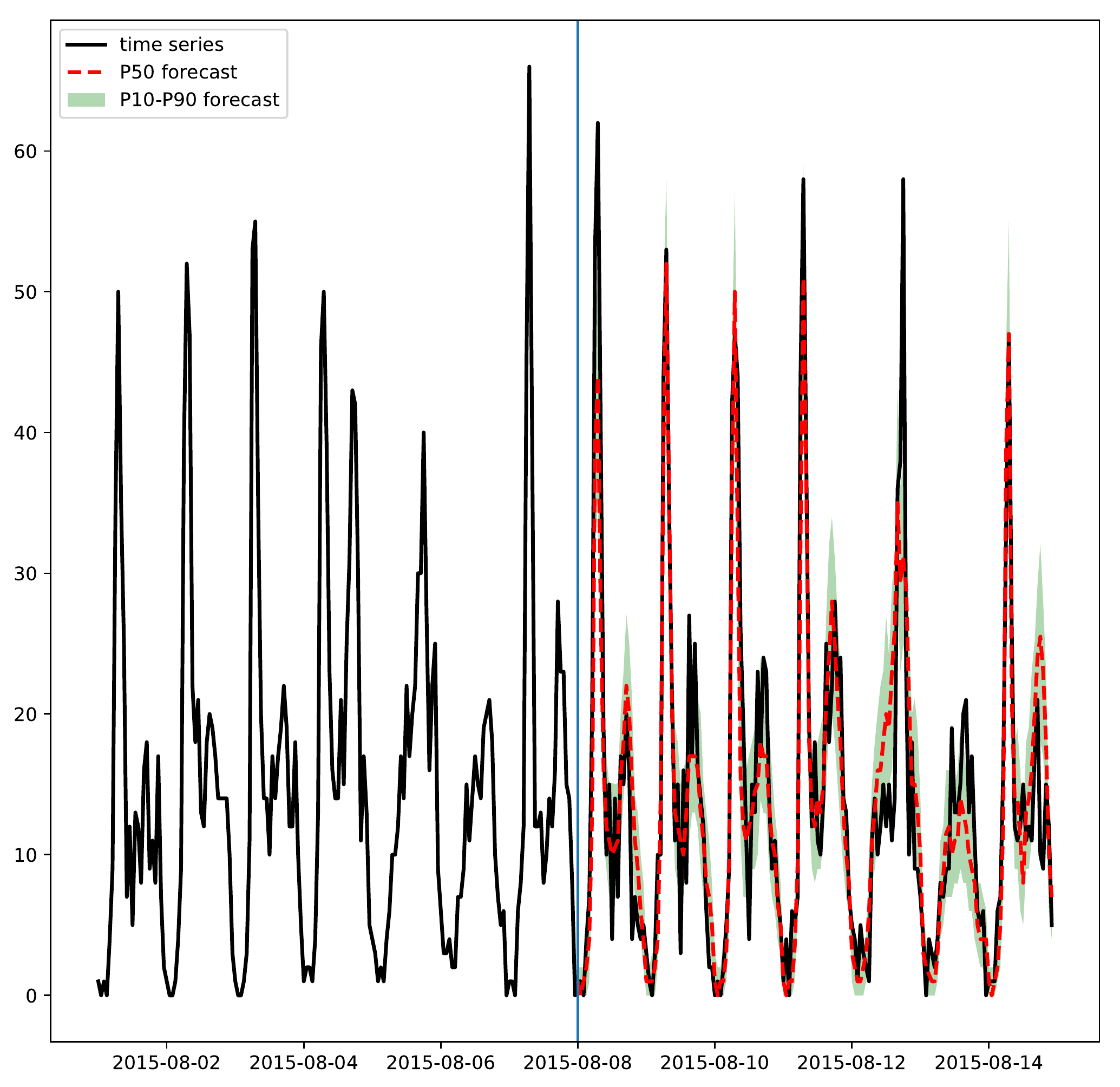} %example_u.png}                
    \end{minipage}  
    \begin{minipage}{.24\textwidth}
        \centering
        \includegraphics[width=\linewidth]{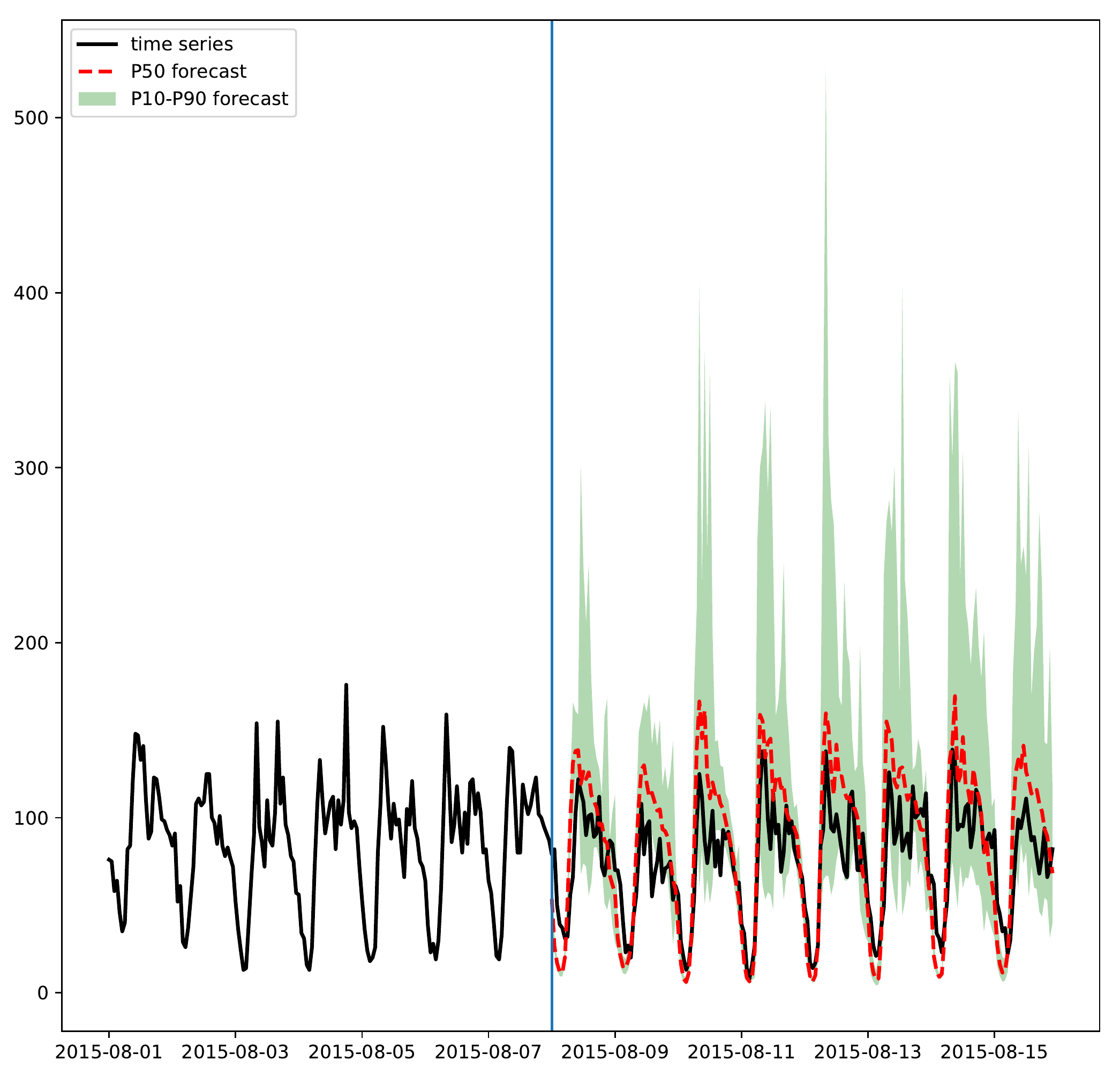} 
    \end{minipage}          
    \begin{minipage}{.24\textwidth}
        \centering
        \includegraphics[width=\linewidth]{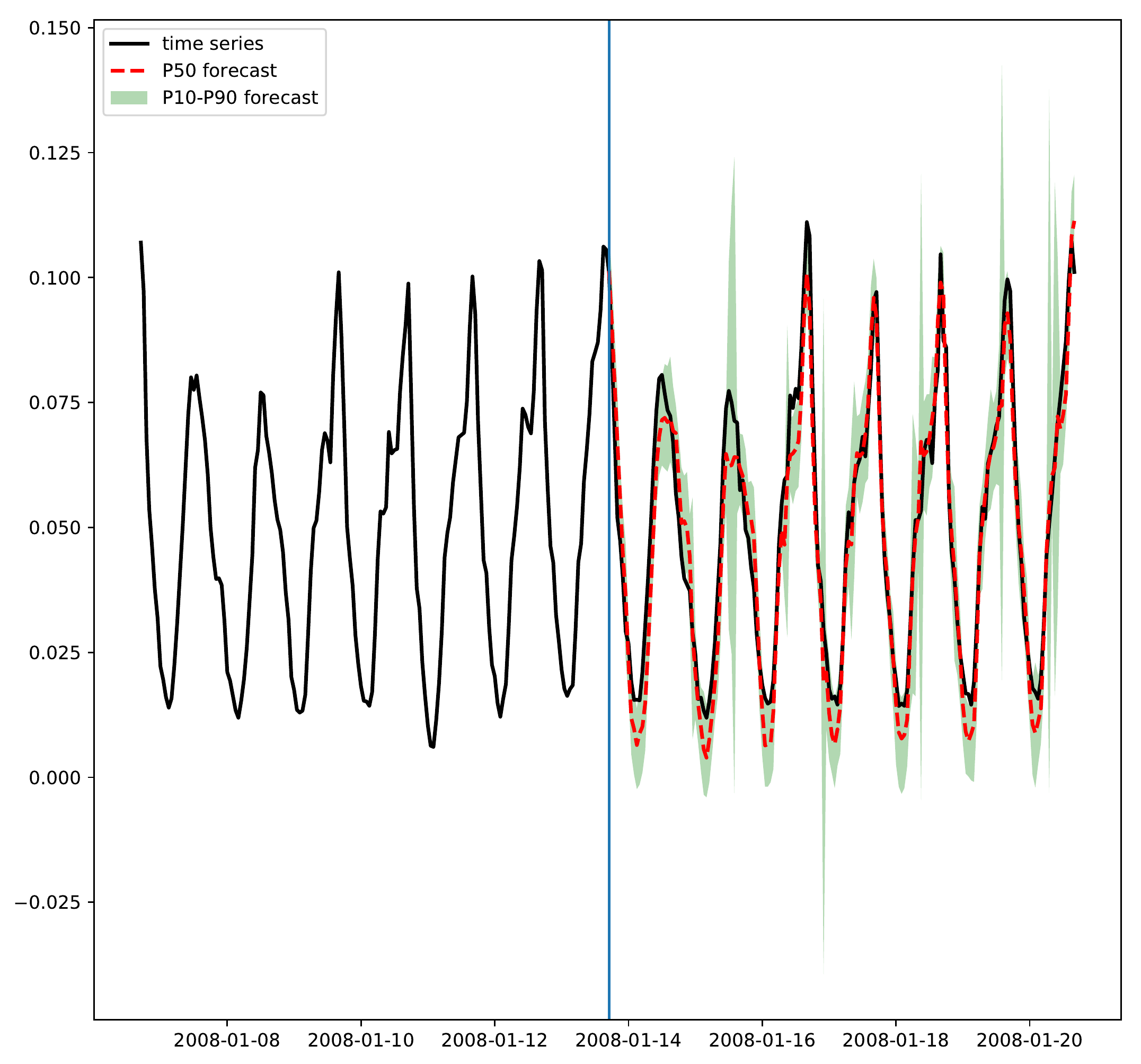} %example_u.png}                
    \end{minipage}    
    \begin{minipage}{.24\textwidth}
        \centering
        \includegraphics[width=\linewidth]{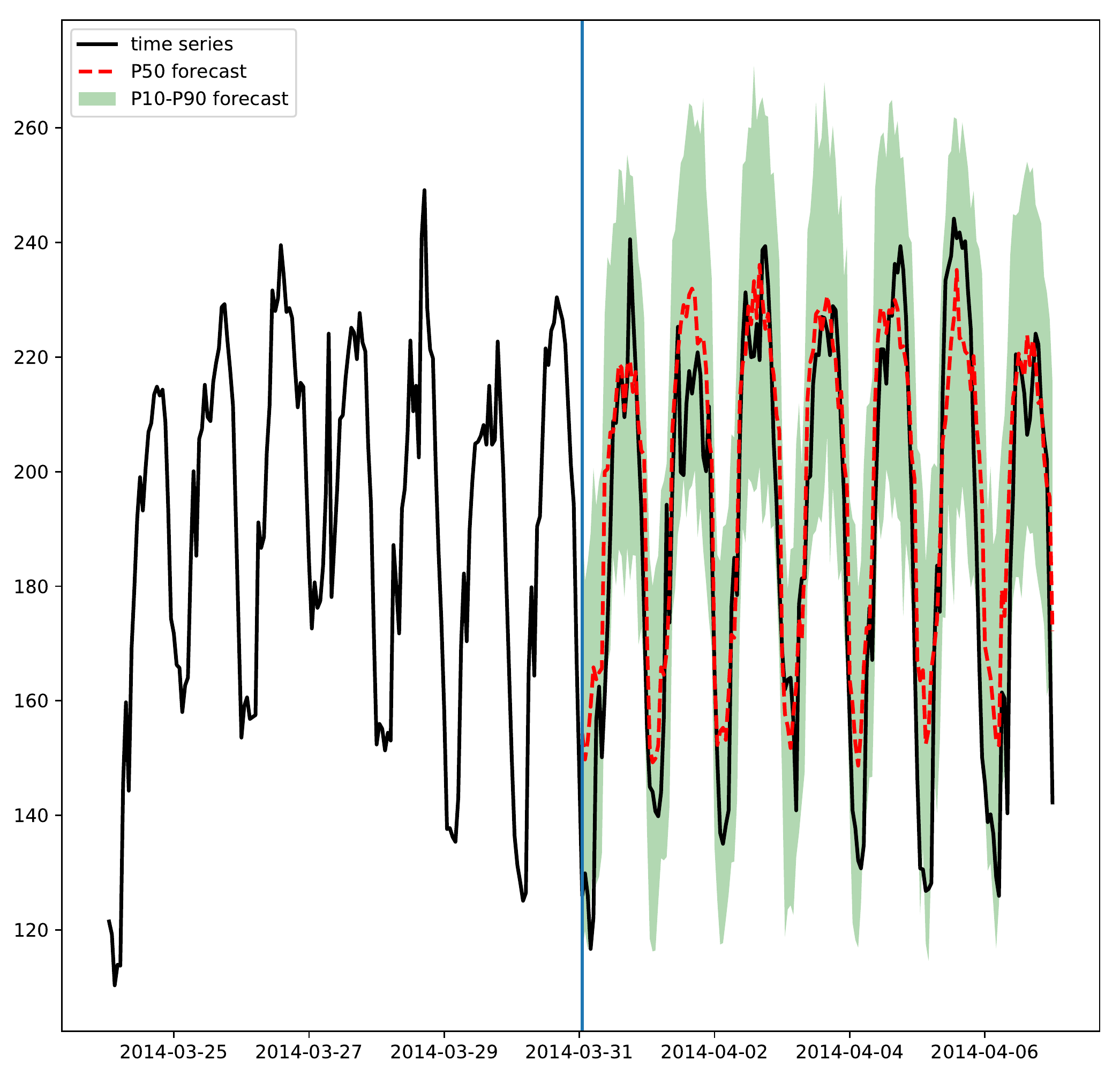}                
    \end{minipage}  
    \caption{Example forecast for (top row) \texttt{uber} (left) and \texttt{taxi} (center left) in the cold-start situation, and regular setting (center right) and \texttt{traffic}. For the \texttt{taxi} cold-start example, our model fails to produce a reasonable forecast, possibly due to insufficient information of corresponding latitude and longitude. Bottom role: another set of example forecasts for \texttt{uber, taxi, traffic} and \texttt{electricity}.}
     \label{fig:example2}
\end{figure}

We provide the result of one trial for Prophet in Table \ref{tab:results} since its classical methods have less variability between experiments and are more computationally expensive.  The acronyms DA, P, MR, DF in these tables stand for DeepAR, Prophet, MQ-RNN and DeepFactor, respectively.  Figure~\ref{fig:example2} illustrates example forecasts from our model. 

In our experiments, we do not tune the hyper-parameters of the methods.  With tuning and a much bigger model, DeepAR would achieve higher accuracy in our preliminary experiments. For example, with the SageMaker HPO, we achieve the DeepAR's best performance of 0.118 MAPE and 0.042 P90 Loss on the \texttt{electricity} dataset for horizon 72. The hyper-parameters are selected to be a two-layer LSTM each with 40 units, a learning rate of 5e-3, 2000 epochs and an early stopping criteria of 200 epochs. As we can see, this metric is on par with our results in Table~\ref{tab:results} and DF achieves such accuracy with much less number of parameters.

\begin{figure}[h]
    \centering
    %\begin{minipage}{.38\textwidth}
       % \centering
        %\includegraphics[width=\linewidth]{figs/MAPE_sub.pdf} 
    %\end{minipage}%
    \begin{minipage}{.33\textwidth}
        \centering
        \includegraphics[width=\linewidth]{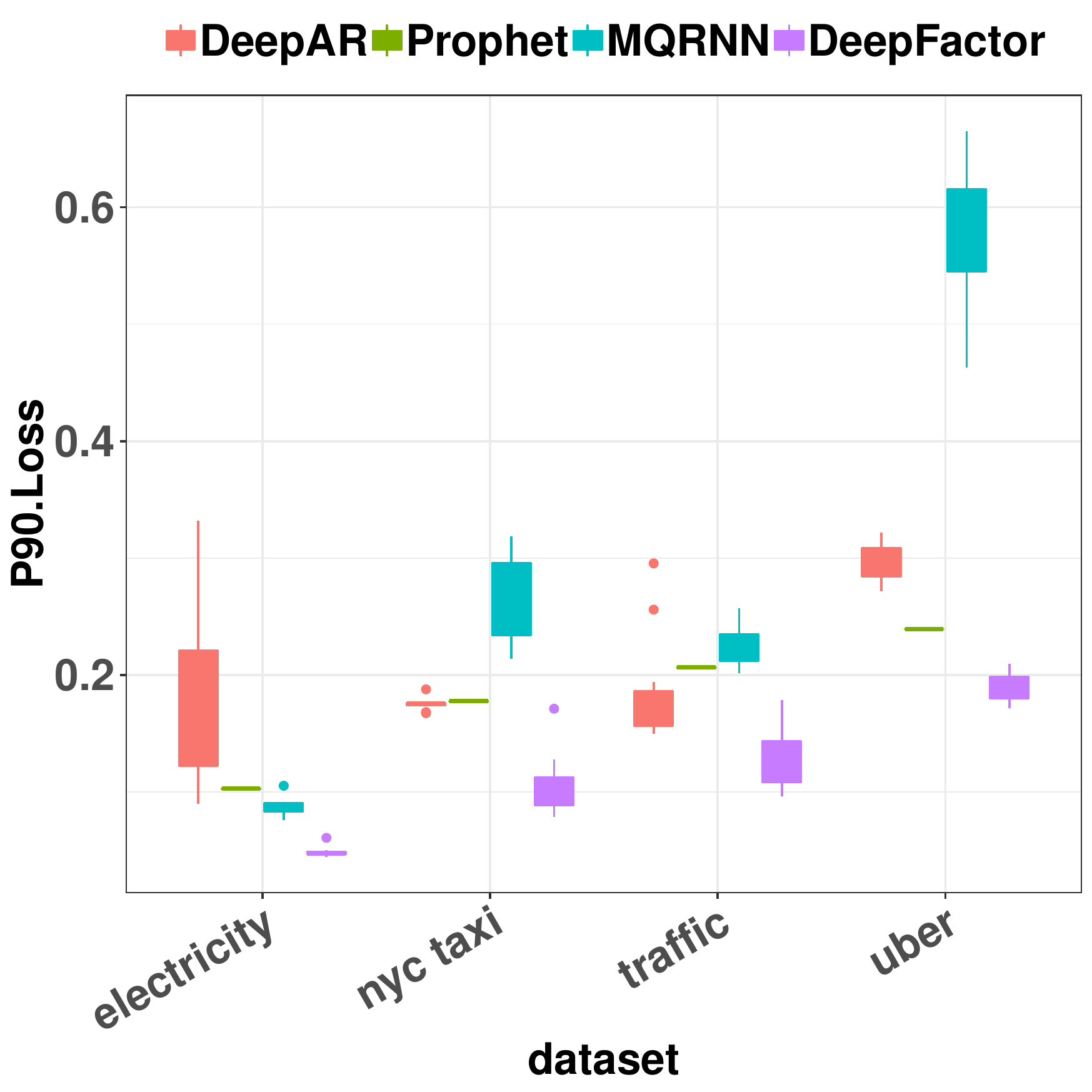}
    \end{minipage}
      \begin{minipage}{.33\textwidth}
        \centering
        \includegraphics[width=\linewidth]{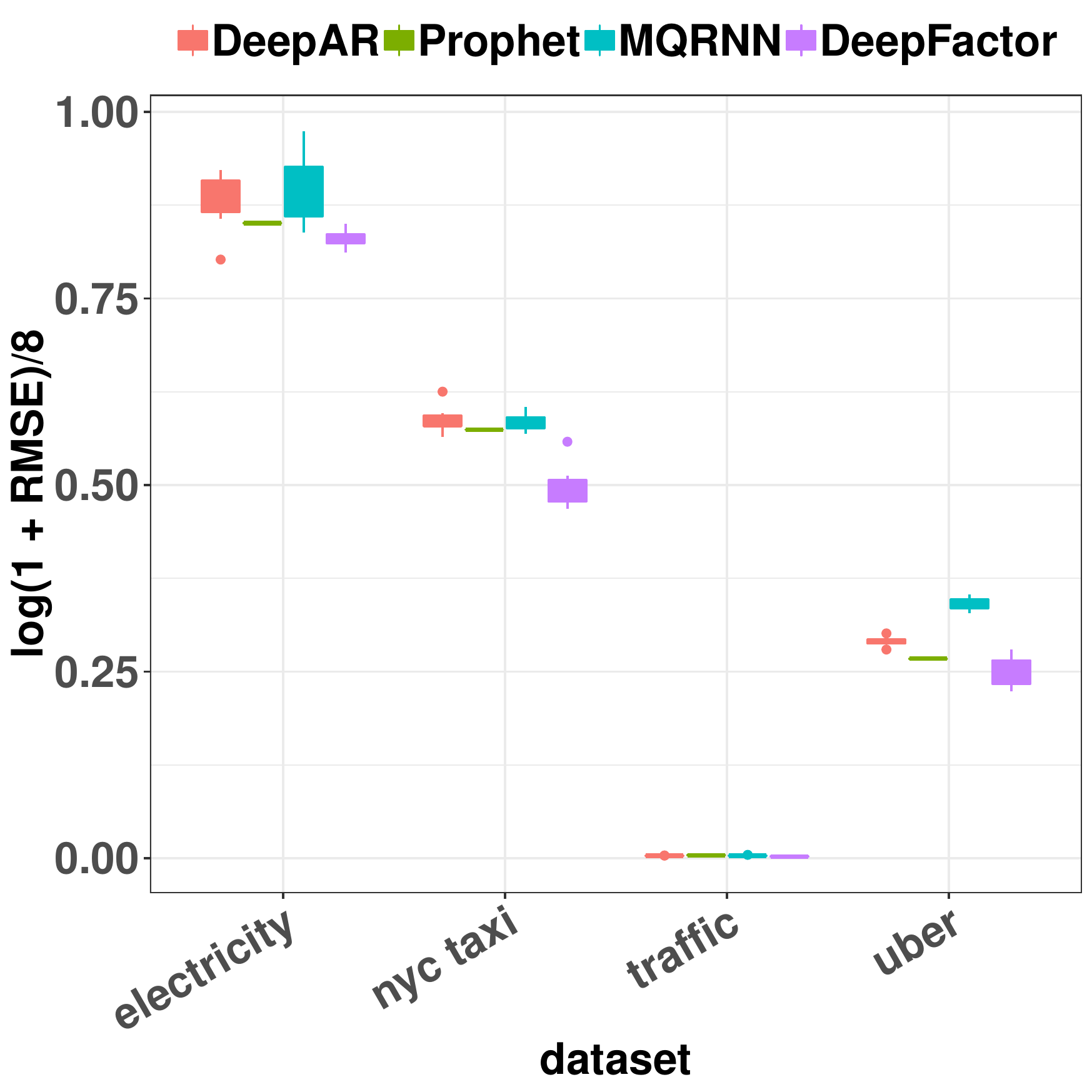}
    \end{minipage}
   % \begin{minipage}{.33\textwidth}
     %  \centering
      %  \includegraphics[width=\linewidth]{figs/RMSE_sub}
  %  \end{minipage}       
    \caption{P90QL and $\log(1+\text{RMSE})$ normalized by 8 results for the forecast horizon 72 in Tables \ref{tab:results} and \ref{tab:results_RMSE}, respectively. Purple denotes the proposed method.}   %Prophet is a widely used statistics algorithm at Facebook, which we use to benchmark against.}
\end{figure}

\begin{figure}[h]
    \centering
    \begin{minipage}{.33\textwidth}
        \centering
        \includegraphics[width=\linewidth]{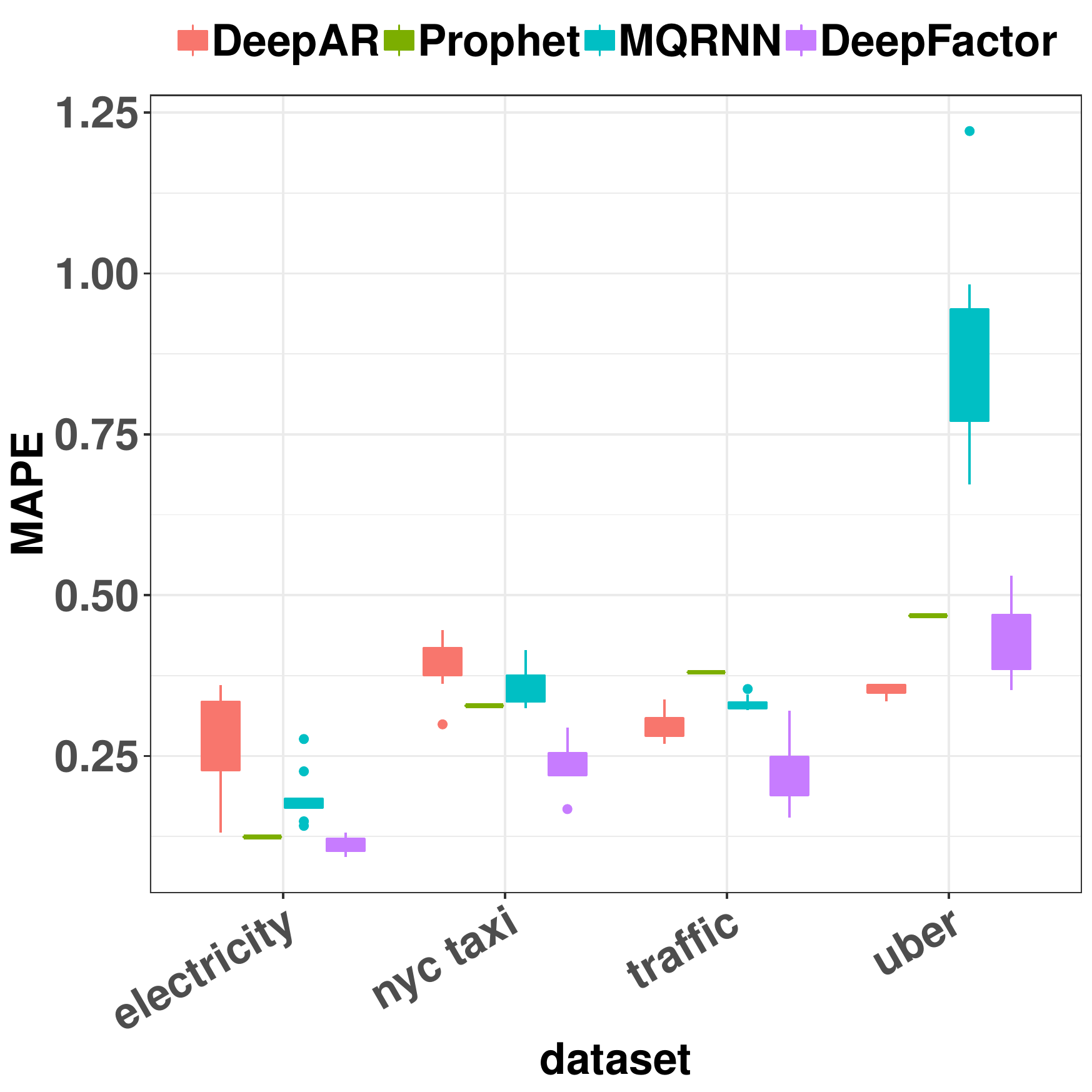} 
    \end{minipage}%
    \begin{minipage}{.33\textwidth}
        \centering
        \includegraphics[width=\linewidth]{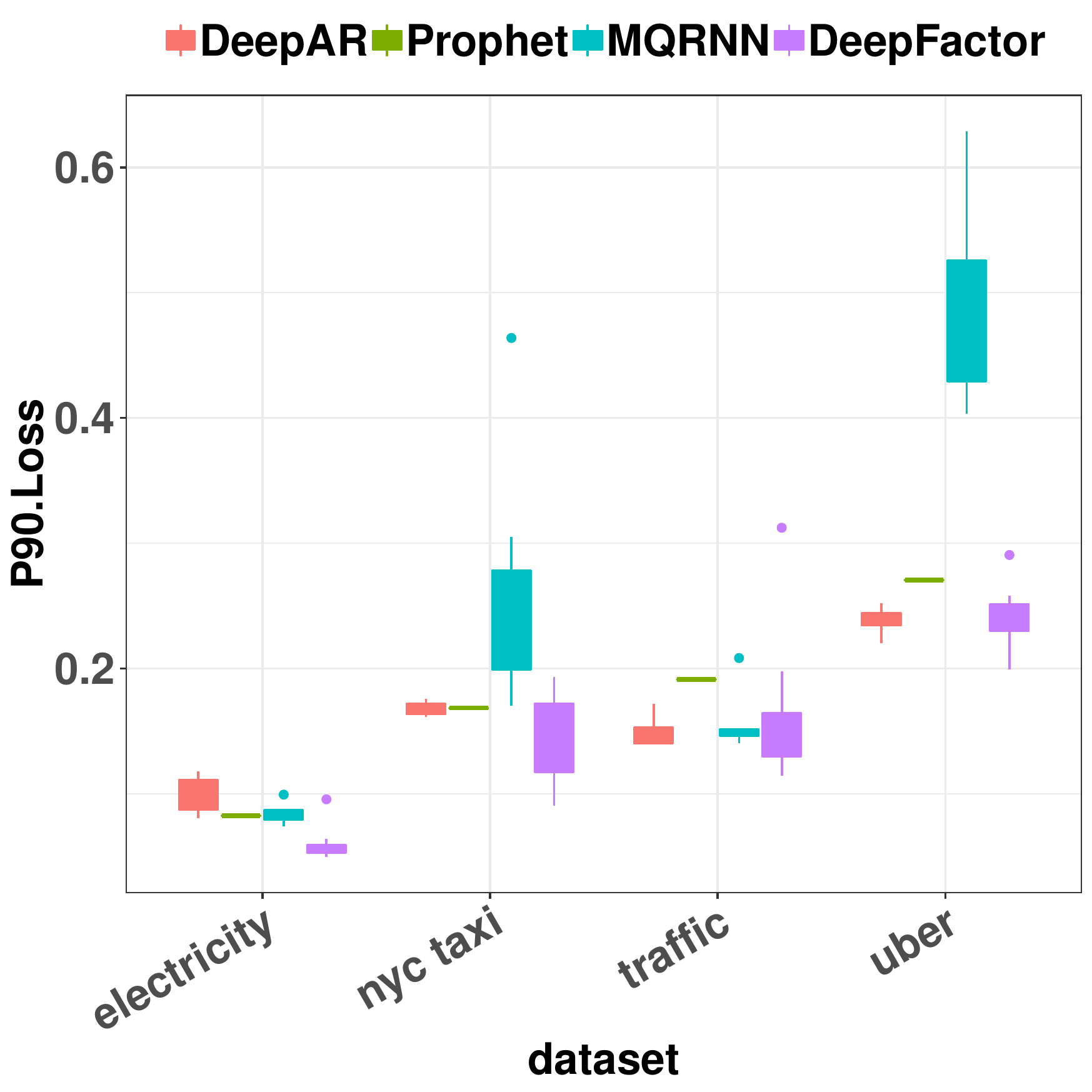}
    \end{minipage}
      \begin{minipage}{.33\textwidth}
        \centering
        \includegraphics[width=\linewidth]{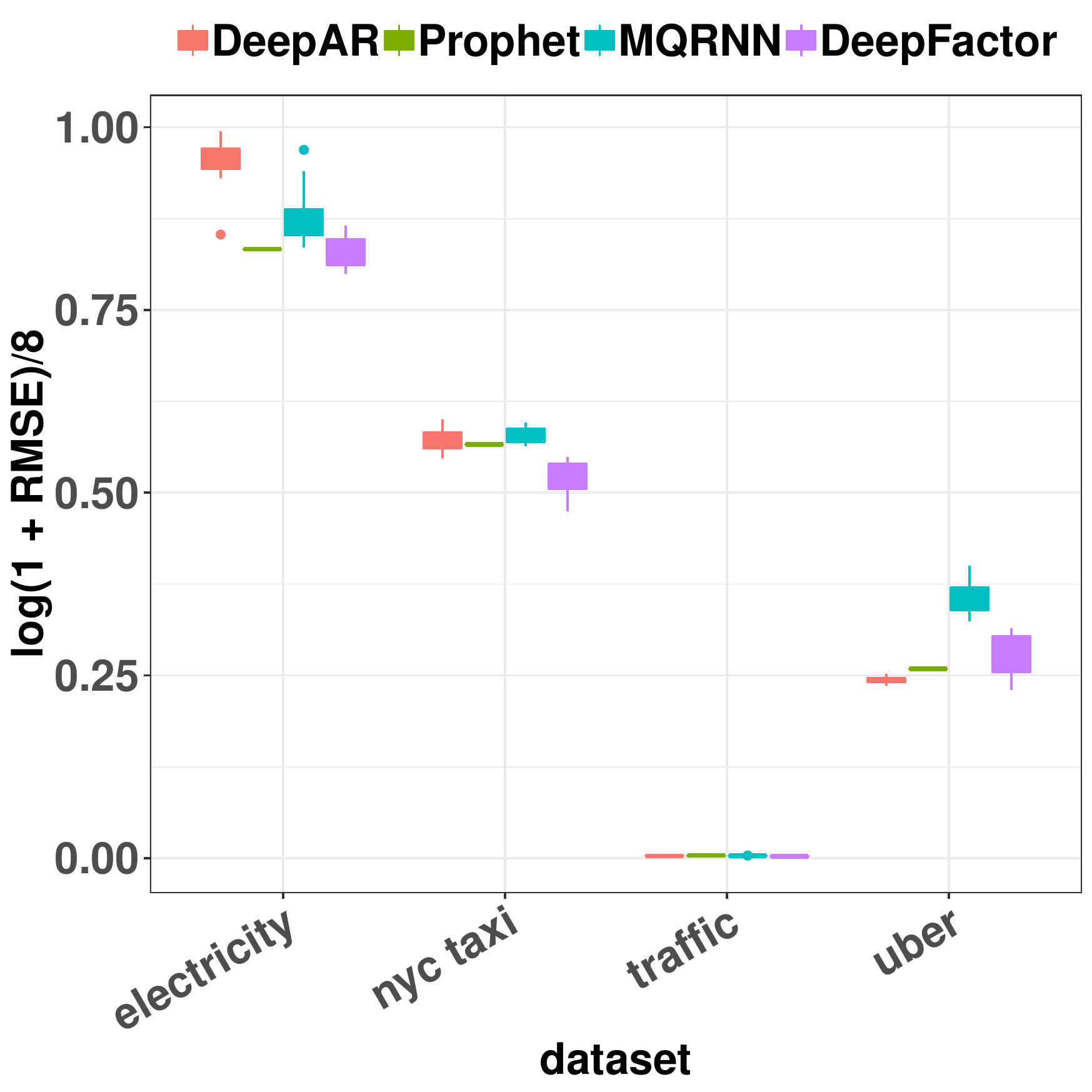}
    \end{minipage}
   % \begin{minipage}{.33\textwidth}
     %  \centering
      %  \includegraphics[width=\linewidth]{figs/RMSE_sub}
  %  \end{minipage}       
    \caption{P50QL (MAPE) P90QL and $\log(1+\text{RMSE}) / 8.0$ (to make them in the same scale) results for forecast horizon 24 in Tables \ref{tab:results} and \ref{tab:results_RMSE}, respectively. Purple denotes the proposed method.}  %Prophet is a widely used statistics algorithm at Facebook, which we use to benchmark against.}
\end{figure}

\begin{table*}[ht]
\small
\centering
%\scriptsize
\begin{tabular}{l|c|cccc|}
\toprule
\multirow{2}{*}{\textsc{ds}} & \multirow{2}{*}{\textsc{h}} & \multicolumn{4}{c}{\textsc{RMSE}}  \\
\cline{3-6}
{} & & DA & P & MR & DF \\
\midrule 
\multirow{2}{*}{\textbf{E}} & 72  & 1194.421$ \pm$ 295.603& 902.724 &1416.992 $\pm$ 510.705 & \textbf{770.437 $\pm$ 68.151} \\
                            & 24  &2100.927 $\pm$ 535.733 & \textbf{783.598} & 1250.233 $\pm$  509.159 & 786.667 $\pm$ 144.459 \\
\midrule
\multirow{2}{*}{\textbf{N}} & 72 & 108.963 $\pm$ 15.905 & 97.706 &106.588 $\pm$ 9.988 & \textbf{53.523 $\pm$ 13.096}	\\
                            & 24 & 98.545 $\pm$  13.561 & 91.573 &102.721 $\pm$ 9.640 & \textbf{63.059 $\pm$ 11.876} \\
\midrule 
\multirow{2}{*}{\textbf{T}} & 72  & 0.027 $\pm$ 0.001 & 0.032  &0.029 $\pm$ 0.003 &\textbf{0.019 $\pm$ 0.002}\\
                            & 24 &0.025 $\pm$ 0.001 & 0.028 &0.028 $\pm$  0.001 & \textbf{0.021 $\pm$ 0.003}\\
\midrule
\multirow{2}{*}{\textbf{U}} & 72 &9.218 $\pm$ 0.486 & 7.488 &14.373 $\pm$  0.997 &\textbf{6.393 $\pm$ 1.185}\\
                            & 24  &\textbf{6.043 $\pm$ 0.295} & 6.948 & 16.585 $\pm$ 3.452 & 8.044 $\pm$ 2.307 \\
\bottomrule
\end{tabular}
\caption{RMSE: Results for short-term (3-day forecast, 72-hour) and near-term (24-hour forecast) scenario with one week of training data.}
\label{tab:results_RMSE}
\end{table*}    

\end{document}